\documentclass{article}

\usepackage{amsmath,amssymb,amsfonts,amstext}%
\usepackage{graphicx}%
\usepackage{multirow}%
\usepackage{import}
\usepackage{algorithm}%
\usepackage{algorithmicx}%
\usepackage{algpseudocode}%
\usepackage[preprint, nonatbib]{neurips_2024}

\usepackage[utf8]{inputenc} % allow utf-8 input
\usepackage[T1]{fontenc}    % use 8-bit T1 fonts
\usepackage[hidelinks]{hyperref}       % hyperlinks
\usepackage{url}            % simple URL typesetting
\usepackage{booktabs}       % professional-quality tables
\usepackage{nicefrac}       % compact symbols for 1/2, etc.
\usepackage{microtype}      % microtypography
\usepackage{xcolor}         % colors
\usepackage{biblatex} %Imports biblatex package
\usepackage{caption}
\title{Online Model-based Anomaly Detection in Multivariate Time Series: Taxonomy, Survey, Research Challenges and Future Directions}

\author{%
  Lucas Correia\\
  Mercedes-Benz AG\\
  Stuttgart\\
  Germany \\
  \texttt{lucas.correia@mercedes-benz.com} \\
  \And
  Jan-Christoph Goos \\
  Mercedes-Benz AG\\
  Stuttgart\\
  Germany \\
  \AND
  Philipp Klein \\
  Mercedes-Benz AG\\
  Stuttgart\\
  Germany \\
  \And
  Thomas Bäck \\
  Leiden University \\
  Leiden \\
  The Netherlands \\
  \And
  Anna V. Kononova \\
  Leiden University \\
  Leiden \\
  The Netherlands \\
}

\newcommand{\lc}[1]{\textcolor{black}{#1}}

\algnewcommand\algorithmicinput{\textbf{Input:}}
\algnewcommand\Input{\item[\algorithmicinput]}

\algnewcommand\algorithmicresult{\textbf{Result:}}
\algnewcommand\Result{\item[\algorithmicresult]}

\begin{document}

\maketitle

\begin{abstract}
Time-series anomaly detection plays an important role in engineering processes, like development, manufacturing and other operations involving dynamic systems.
These processes can greatly benefit from advances in the field, as state-of-the-art approaches may aid in cases involving, for example, highly dimensional data.
\lc{To provide the reader with understanding of the terminology, this survey introduces a novel taxonomy where a distinction between online and offline, and training and inference is made.
Additionally, it presents the most popular data sets and evaluation metrics used in the literature, as well as a detailed analysis.
Furthermore, this survey provides an extensive overview of the state-of-the-art model-based online semi- and unsupervised anomaly detection approaches for multivariate time-series data, categorising them into different model families and other properties. 
The biggest research challenge revolves around benchmarking, as currently there is no reliable way to compare different approaches against one another. 
This problem is two-fold: on the one hand, public data sets suffers from at least one fundamental flaw, while on the other hand, there is a lack of intuitive and representative evaluation metrics in the field.
Moreover, the way most publications choose a detection threshold disregards real-world conditions, which hinders the application in the real world. 
To allow for tangible advances in the field, these issues must be addressed in future work.}
\end{abstract}

%% List of abbreviations
\begin{table}[h!]
\centering
\begin{tabular}{ll} \hline
Acronym & Full Designation \\ \hline\hline
VAE     & Variational Autoencoder                 \\
GAN     & Generative Adversarial Network                 \\
DL      & Deep Learning                 \\
MIT-BIH & Massachusetts Institute of Technology - Beth Israel Hospital \\
HRSS    & High Rack Storage System                \\
TEP     & Tennessee-Eastman Process                \\
WADI    & Water Distribution                 \\
SWaT    & Secure Water Treatment                 \\
SMAP    & Soil Moisture Active Passive                 \\
MSL     & Mars Science Laboratory                 \\
CNC     & Computer Numerical Control                 \\
SMD     & Server Machine Data                  \\
MSDS    & Multi-Source Distributed System                 \\
PSM     & Pooled Server Metrics                 \\
LSTM    & Long Short-term Memory                 \\
GRU     & Gated Recurrent Unit                 \\
CNN     & Convolutional Neural Network                 \\
TCN     & Temporal Convolutional Network \\
POT     & Peaks over Threshold                 \\ \hline
\end{tabular}
\caption*{\lc{List of acronyms used in this publication.}}
\end{table}
%% \linenumbers

\section{Introduction}\label{sec:introduction}
As a result of the fourth industrial revolution, also known as industry 4.0, immense amounts of data are collected from sensors mounted at different checkpoints in many processes in research and development, manufacturing and testing \cite{xu_internet_2014}. 
This data can expose subtle but important trends and correlations, as well as give the data user key insights on how to optimise engineering systems and processes, which can potentially provide a company with a competitive advantage in the market.
Recording high-quality data is important, as incomplete or anomalous data can negate any potential benefits that can be extracted from it. 
With the rise of industry 4.0, anomaly detection has therefore gained relevance over the past decade, with the bar being set ever higher as data becomes more and more high dimensional \cite{zimek_survey_2012, erfani_high-dimensional_2016}. 
\lc{Time-series data analysis in particular has been an active research area due to the ubiquity of dynamic systems in industry.
Dynamic systems are systems that can be described with features that are functions of time. Measuring multiple features of such systems, therefore, yields multivariate time-series data, which then amounts to a representation of the dynamic system. Examples of dynamic systems include aeroplanes, cars or even humans.
This type of data} has gained special attention, given the correlation present between different features and along its time axis, which allows time series to represent complex dynamic processes.
Thus, reliably finding anomalies in high-dimensional time series is a very active research area.
Approaches related to online applicability are of interest for industrial use, as finding anomalies in a timely manner can help reduce costs and increase operating efficiency.
Deep learning is also a very active research area, owing to increasing computing power and the availability of \lc{large amounts of} data. 
It can be applied to anomaly detection, especially in high dimensional data, which is where traditional approaches have started to reach their limits \cite{chalapathy_deep_2019}.
\lc{In the real world, large amounts of labelled data are often not available, hence why most literature in the anomaly detection field is dedicated to semi-supervised and unsupervised anomaly detection methods, rather than supervised ones. Semi-supervised anomaly detection is defined as a problem where the data set contains only nominal, i.e.\ anomaly-free, sequences, whereas, in the case of unsupervised anomaly detection, the data set is entirely unlabelled \cite{chandola_anomaly_2012}. 
In this work, the word \emph{nominal} is used as a synonym for \emph{normal}, to avoid confusion with a normal distribution.
Semi-supervised anomaly detection can be considered a more relaxed version of unsupervised anomaly detection, as any unsupervised approaches can be applied to a semi-supervised problem \cite{aggarwal_time_2017}.}

Based on these key focus points, the survey is structured as follows: first, a novel taxonomy (Section \ref{sec:taxonomy}) is defined, including anomaly types, approaches to anomaly detection and the various cases that are encompassed in the online anomaly detection domain.
Then, work related to this publication (Section \ref{sec:related_work}) is presented. 
This includes surveys that specialise in time-series anomaly detection and feature, at least partially, content on online detection or detection in streaming data. 
Following that, a section introduces benchmarking aspects of time-series anomaly detection.
It discusses publicly available data sets, as well as the anomaly detection metrics (Section \ref{sec:benchmark}) used in the literature to evaluate approaches. 
Sections \ref{sec:predictive_modelling} - \ref{sec:transformer_modelling} discuss the approaches in the literature used to detect anomalies in multivariate time series. 
The publications are grouped by architectural similarity and sorted by their ability to learn and infer in an online manner, as well as publishing date. 
\lc{Additionally, it is specified whether they can detect anomalies without any labelled data.
The purpose of this label is to distinguish approaches that set a threshold following certain assumptions and approaches that set a threshold which maximises anomaly detection performance by means of labelled data.}
Section \ref{sec:discussion} presents a discussion around the surveyed model types, as well as an assessment of what research gaps exist and what areas deserve to be further looked into. 
\lc{After that, Section \ref{sec:applications} is dedicated to outlining real-world applications of online model-based anomaly detection.}
The conclusion (Section \ref{sec:conclusion}) summarises the current state of the time-series anomaly detection landscape, while also concisely highlighting current problems and research gaps. \lc{Managerial insights are provided to bridge the gap between academic research and users in real-world engineering applications.}
\lc{Lastly, Section \ref{sec:outlook} provides a concise outline of future directions that should be taken to address the issues pointed out in the survey.}
\begin{figure}[h!]
    \centering
    \def\svgwidth{\textwidth}
    \scriptsize
    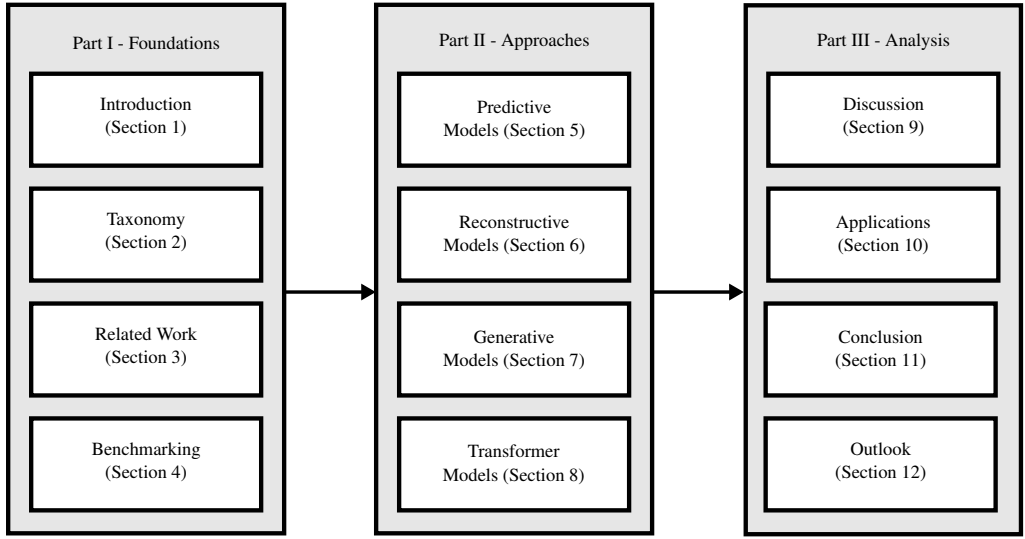
    \caption{A visual representation of the structure of the survey.}
    \label{fig:contents}
\end{figure}
For a visual representation of the structure of the survey, please refer to Figure \ref{fig:contents}.

\section{Taxonomy in Time-series Anomaly Detection}\label{sec:taxonomy}
\begin{figure}[h!]
    \centering
    \def\svgwidth{0.80\textwidth}
    \scriptsize
    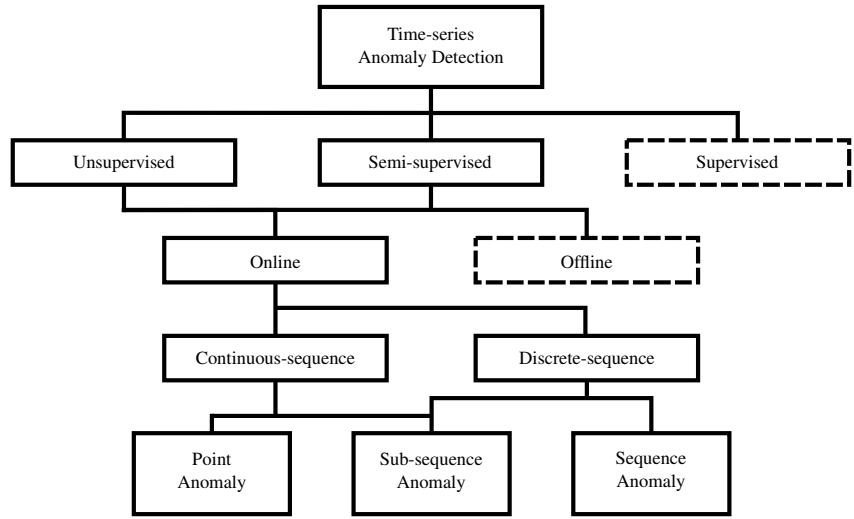
    \caption{\lc{A visual representation of taxonomy used in this survey.}}
    \label{fig:taxonomy}
\end{figure}
\lc{To provide the reader with an illustration of the relationship between terms introduced in the taxonomy used an overview is represented graphically in Figure \ref{fig:taxonomy}.}

\subsection{Anomaly Types and Detection} \label{subsec:anomaly_types}
Anomalies in time series can be assumed to have different shapes and forms.
A commonly used and accepted definition \cite{hawkins_identification_1980} reads as follows: 
\begin{center} \textit{"An observation which deviates so much from other observations as to arouse suspicions that it was generated by a different mechanism."} \end{center}
Over the years, several taxonomies have been proposed to better classify different anomalies. 
This work mostly follows the taxonomy suggested by \cite{blazquez-garcia_review_2021}, where anomalies are classified into point outliers, sub-sequence outliers (also known as collective anomalies in the literature) and outlier time series. 
In this work, these three types will be from now on referred to as \textit{point anomalies}, \textit{sub-sequence anomalies} and \textit{whole-sequence anomalies}, respectively.

A \textit{point anomaly} is defined as a value at a time step that does not conform to the typical behaviour of a system. 
Consider a testing subset $\mathcal{D}^{\text{test}}$ in data set $\mathcal{D}$ containing $N$ sequences, such that $\mathcal{D}^{\text{test}} = [\mathcal{S}_1, ..., \mathcal{S}_n, ..., \mathcal{S}_{N}]$, where $\mathcal{S}_n \in \mathbb{R}^{T_{n} \times d_\mathcal{D}}$.
In a given multivariate sequence $\mathcal{S}_n$, an anomaly event $\mathcal{A} \in \mathbb{R}^{S \times d_\mathcal{A}}$ of length $S$ that is detected in $d_\mathcal{A}$ channels, where $d_\mathcal{A}\leq d_\mathcal{D}$, is considered a point anomaly when $S=1$. 
\begin{figure}[h!]
    \centering
    \includegraphics[width=\textwidth]{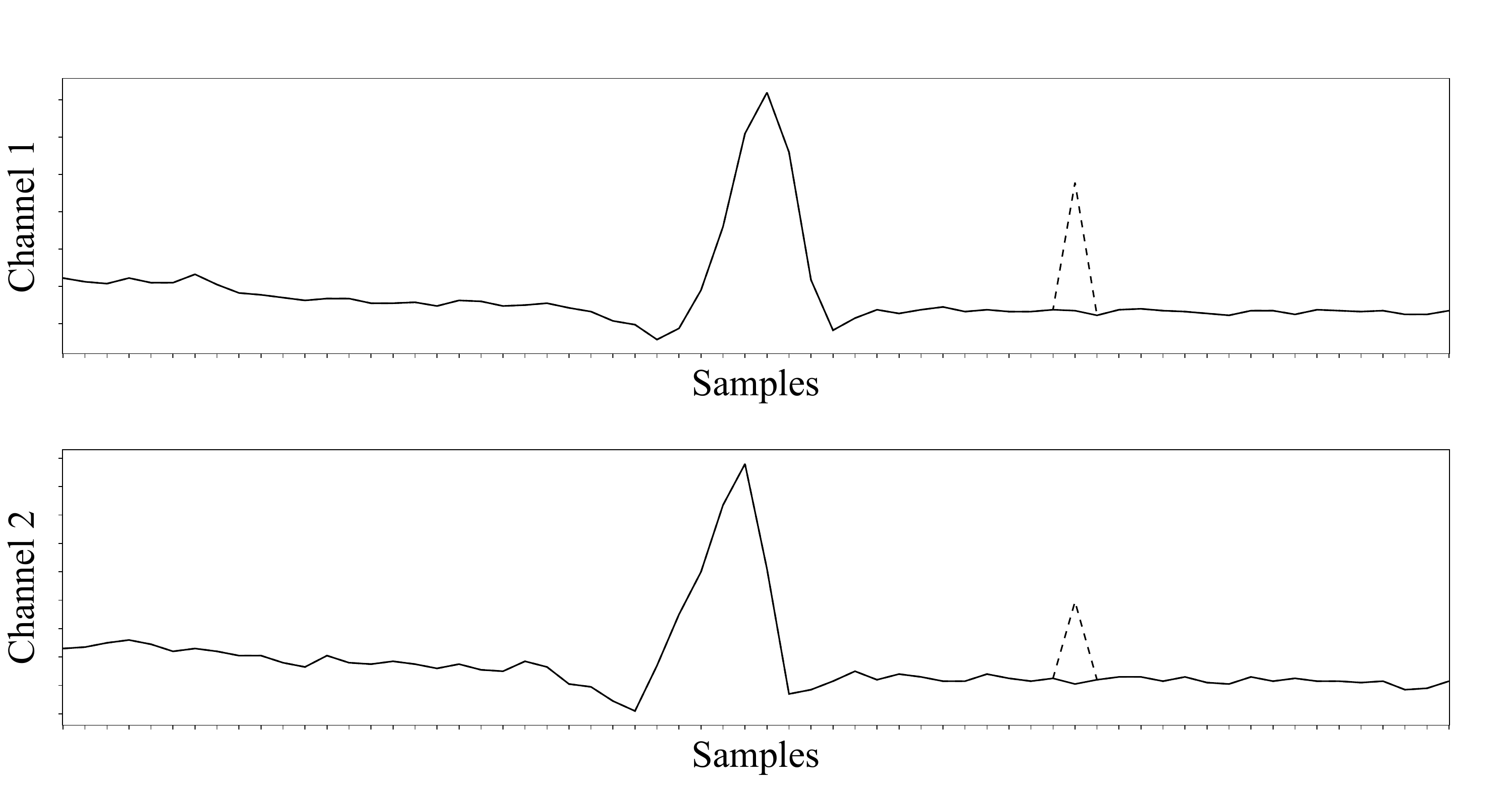}
    \caption{Example of a \emph{point anomaly} in both channels. The continuous line represents the ground-truth time series and the traced line the anomaly. The ground-truth time series shown is a heartbeat from the MIT-BIH Arrhythmia Database \cite{moody_impact_2001}.}
    \label{fig:point_anomaly_2}
\end{figure}
An example of a point anomaly is illustrated in Figure \ref{fig:point_anomaly_2}. 
While more easily detectable than the other types, these anomalies are rare events in the real world, as systems affected cannot usually return to a nominal state before the next time step arrives unless the sampling rate is very low.

\textit{Sub-sequence anomalies} are defined by a series of anomalous time steps, i.e.\ a sub-sequence within a time series that does not reflect the nominal behaviour of a system. 
In a given multivariate sequence $\mathcal{S}_n$, an anomaly event $\mathcal{A} \in \mathbb{R}^{S \times d_\mathcal{A}}$ of length $S$ that is detected in $d_\mathcal{A}$ channels, where $d_\mathcal{A}\leq d_\mathcal{D}$, is considered a sub-sequence anomaly when $1<S<T$. 
In a real-world system, this type of anomaly may occur when a component in the said system runs at reduced functionality or fails but manages to recover to a nominal state after a period of time.
\begin{figure}[h!]
    \centering
    \includegraphics[width=\textwidth]{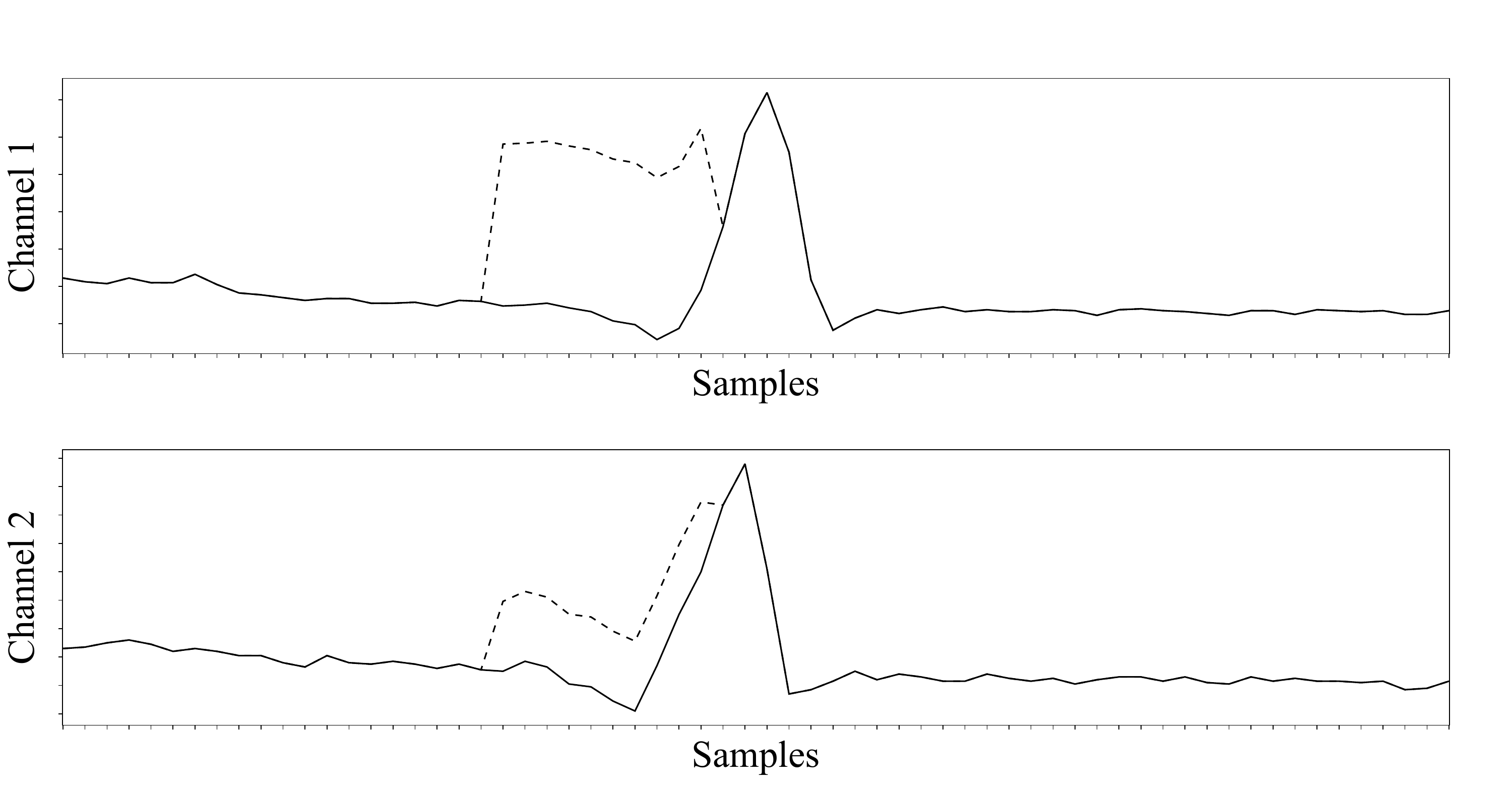}
    \caption{Example of a \emph{sub-sequence anomaly} in both channels. The continuous line represents the ground-truth time series and the traced line the anomaly. The ground-truth time series shown is a heartbeat from the MIT-BIH Arrhythmia Database \cite{moody_impact_2001}.}
    \label{fig:sub_anomaly_2}
\end{figure}
An example of a sub-sequence anomaly is illustrated in Figure \ref{fig:sub_anomaly_2}.

A \textit{whole-sequence anomaly} can be seen as a long sub-sequence that has the same length as the entire sequence.
Hence, in a given multivariate sequence $\mathcal{S}_n$, an anomaly event $\mathcal{A} \in \mathbb{R}^{S \times d_\mathcal{A}}$ of length $S$ that is detected in $d_\mathcal{A}$ channels, where $d_\mathcal{A}\leq d_\mathcal{D}$, is considered a whole-sequence anomaly when $S=T$. 
This type of anomaly can occur when an initial parameter or state deviates from the norm, leading to all observations in the sequence being anomalous as well.
\begin{figure}[h!]
    \centering
    \includegraphics[width=\textwidth]{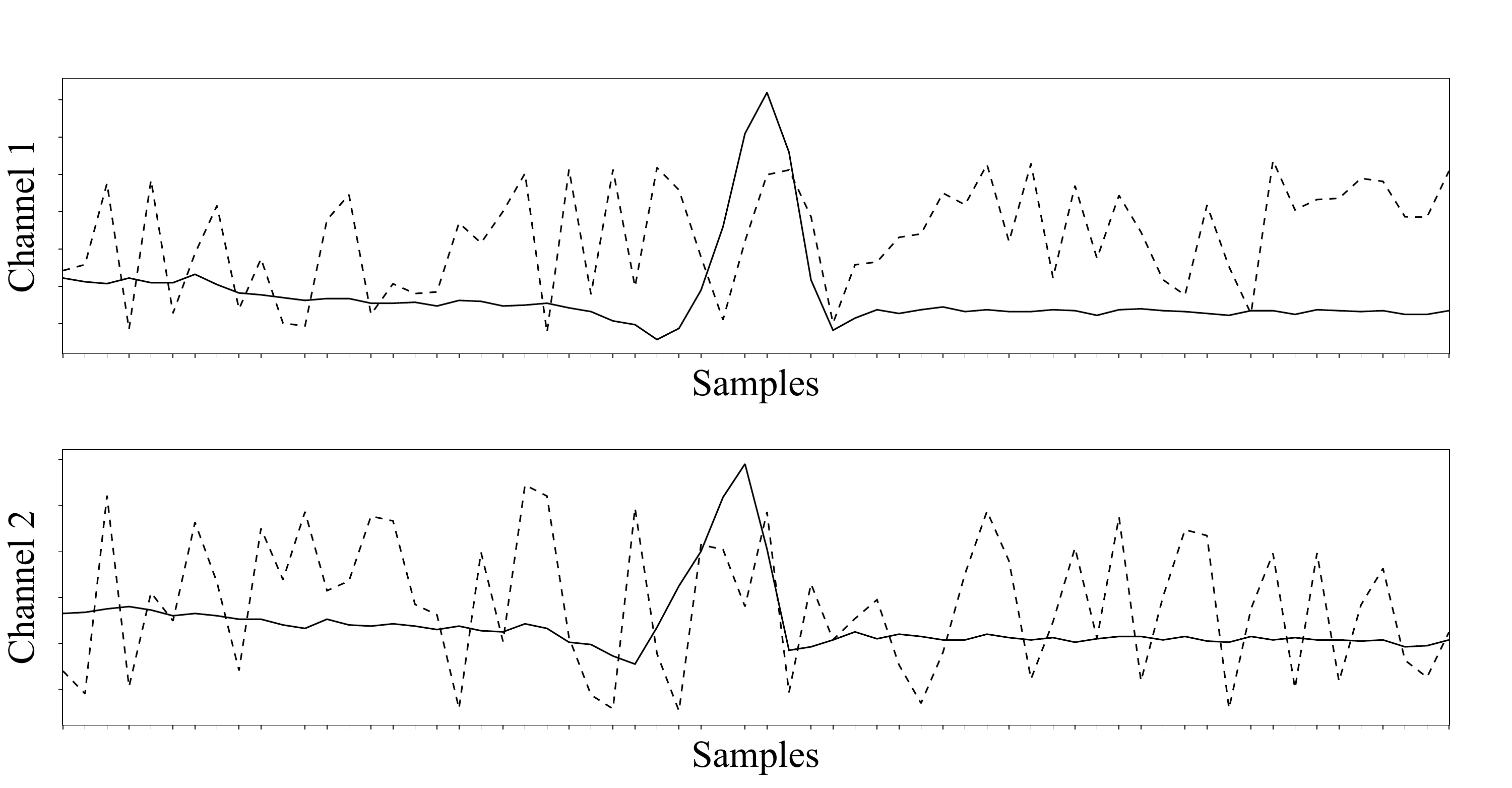}
    \caption{Example of a \emph{whole-sequence anomaly} in both channels. The continuous line represents the ground-truth time series and the traced line the anomaly. The ground-truth time series shown is a heartbeat from the MIT-BIH Arrhythmia Database \cite{moody_impact_2001}.}
    \label{fig:ts_anomaly_2}
\end{figure}
An example of a whole-sequence anomaly is illustrated in Figure \ref{fig:ts_anomaly_2}.

\subsection{Continuous- and Discrete-sequence Anomaly Detection} \label{subsec:continuous_discrete_sequence}
Time-series anomaly detection can be split into two main areas: continuous- and discrete-sequence.
Continuous-sequence anomaly detection is the most common type present in public data sets and is defined as detecting anomalies in a process that exists for a longer continuous time period without breaks, i.e.\ $N=1$ in a testing subset $\mathcal{D}^{\text{test}}$.
This includes monitoring applications like in water distribution, server machines or even heartbeat rhythms.
Use cases of continuous-sequence anomaly detection tend to consist of a singular longer time series which contains nominal and anomalous sub-sequences within it, therefore they can only contain point and sub-sequence anomalies.

Discrete-sequence anomaly detection, in contrast, is defined as detecting anomalies in chunks of processes that happen independently of each other, i.e.\ $N<1$ in testing subset $\mathcal{D}^{\text{test}}$.
One example of a discrete-sequence problem is automotive test benches, where several tests may occur one after another but are not temporally contiguous and hence provide a separate time series for each test.
Here, the testees are not monitored over a longer period of time, but rather the measured time series are evaluated as they are being recorded.
Another example could include quality control of manufactured goods, where samples are randomly taken and tested.
Therefore, data sets for discrete-sequence anomaly detection consist of several nominal and anomalous time series, where a given anomalous time series may be entirely anomalous or only partly.

\lc{\subsection{Online Training and Inference} \label{subsec:online_offline}
Using model-based approaches generally involves two processes: training and inference.
Training is the process of automatically adjusting model parameters by means of optimisation using training data, whereas inference refers to the application, where model parameters are no longer adjusted.
Online approaches are defined as models that infer in an online manner and, optionally, are trained in an online manner.
Online training, therefore, refers to the process of adjusting model parameters as training data is streamed. Unlike offline training, which is done once in most cases, online training runs continuously and for an undefined amount of time. 
Online inference refers to the ability of an approach to detect anomalous behaviour in time steps as soon as they are observed, rather than waiting for the time series to have finished streaming and only then evaluating the entire sequence, i.e.\ offline.
This is especially useful in real-world use cases where timely detection is important.
A more strict version of online anomaly detection is real-time anomaly detection, where the current time step is evaluated before the next one arrives.}

\section{Related Work}
\label{sec:related_work}
Various anomaly detection surveys with a focus on time series and online functionality have been published over the years. 
A summary of the surveys is given in Table \ref{tab:survey_table}.
\lc{The table shows whether the surveys discuss anomaly detection taxonomy, available data sets and metrics, and what type of approaches are identified.}
The list of publications, surveys, and reviews has been found using the keywords \emph{anomaly detection}, \emph{outlier detection}, \emph{unsupervised}, \emph{semi-supervised}, \emph{multivariate}, \emph{time series}, \emph{online}, \emph{real-time} and \emph{streaming data} using a variety of libraries, such as \lc{Institute of Electrical and Electronic Engineers (IEEE)} Xplore, \lc{Elsevier} ScienceDirect, Springer \lc{Link} and the \lc{Association for Computing Machinery (ACM)} Digital Library, in the time between 2000 and 2024.
Up until recently, surveys published on online time-series anomaly detection only discussed conventional methods, which for the purpose of this survey will denote methods not based on data-driven modelling. 
Note that the terminology "time series" and "sequence" are used synonymously throughout this survey.
\lc{To help the reader to more easily distinguish between the terms introduced in the taxonomy in this work and other terms used in related work, the terminology stemming from this work's taxonomy is \emph{emphasised} and hyperlinked to the relevant subsection in Section \ref{sec:taxonomy}.}
\begin{table}[h!]
\centering
\begin{tabular}{lcccccccc}\hline
Publications                        & TX         & DS         & EM         & AB         & SSAD         & USAD         & DL         \\\hline\hline%& ST         \\\hline\hline
Chandola et al. \cite{chandola_anomaly_2012}       & \checkmark &            &            &            &\checkmark  & \checkmark &            \\%& \checkmark \\
Gupta et al. \cite{gupta_outlier_2014}          & \checkmark &            &            &            &            & \checkmark &            \\%& \checkmark \\
Aggarwal \cite{aggarwal_time_2017}          & \checkmark &            &            &            & \checkmark & \checkmark &            \\%& \checkmark \\
Mason et al. \cite{mason_online_2019}           &            &            &     	   &            & \checkmark & \checkmark &            \\%& \checkmark \\
Munir et al. \cite{munir_comparative_2019}      &            &            &            &            & \checkmark & \checkmark & \checkmark \\%& \checkmark \\
Blázquez-García et al. \cite{blazquez-garcia_review_2021} & \checkmark &            &            &            &            & \checkmark & \checkmark \\%& \checkmark \\
Lindemann et al. \cite{lindemann_survey_2021}       & \checkmark &            &            &            & \checkmark & \checkmark & \checkmark \\%&            \\
Ngo Bibinbe et al. \cite{ngo_bibinbe_survey_2022}     &            &            &            &            &            & \checkmark & \checkmark \\%& \checkmark \\
Sgueglia et al. \cite{sgueglia_systematic_2022}    & \checkmark & \checkmark &            &            &            & \checkmark & \checkmark \\%& \checkmark \\
Li and Jung \cite{li_deep_2023}                & \checkmark & \checkmark &            &            & \checkmark & \checkmark & \checkmark \\%&            \\ 
\textbf{Ours}                       & \checkmark & \checkmark & \checkmark & \checkmark & \checkmark & \checkmark & \checkmark \\\hline%&            \\ \hline
\end{tabular}
\caption{Summary of surveys discussing online anomaly detection in time series. Key is as follows, TX: anomaly detection taxonomy, DS: data set overview, EM: evaluation metrics overview, AB: analysis of benchmarking, SSAD: semi-supervised anomaly detection, USAD: unsupervised anomaly detection, DL: deep learning approaches.}
\label{tab:survey_table}
\end{table}

\lc{According to \cite{chandola_anomaly_2012}, there are two relevant problem formulations in time-series anomaly detection: sequence-based and contiguous sub-sequence-based anomaly detection.
They are somewhat analogous to \hyperref[subsec:continuous_discrete_sequence]{\emph{discrete-sequence and continuous-sequence anomaly detection}}, though associated with fairly rigid definitions. 
In their work, \cite{chandola_anomaly_2012} frame sequence-based and contiguous sub-sequence-based anomaly detection such that it can only feature whole-sequence anomalies and sub-sequence anomalies, respectively, which may not be the case in the real world.
Furthermore, they dedicate a section discussing online anomaly detection in the context of \hyperref[subsec:continuous_discrete_sequence]{\emph{discrete-sequence and continuous-sequence}} problem framing.}

\lc{\cite{gupta_outlier_2014} categorise time-series anomaly detection problems into five different fields, with the two relevant being so-called time-series data and stream data, essentially designating offline and online anomaly detection.
Time-series data is further split into two subcategories: single time series, where exactly one time series is evaluated and time-series databases, where several time series are evaluated.
In a single time series, \hyperref[subsec:anomaly_types]{\emph{point and sub-sequence anomalies}} can be detected, while in the case of time-series databases, \hyperref[subsec:anomaly_types]{\emph{sub-sequence and whole-sequence anomalies}} can be found, though it is not mentioned why a point anomaly cannot be detected in a time series database.
Additionally, a special case exists where the database can be compared to a single test sequence.
Single time series and time-series databases are analogous to \hyperref[subsec:continuous_discrete_sequence]{\emph{discrete-sequence and continuous-sequence anomaly detection}}, though here they are formulated such that only a testing set is available, an assumption not made in the taxonomy proposed in Subsection \ref{subsec:continuous_discrete_sequence}. 
Also, for the case where a time-series database is compared with a test time series, it is assumed that the test time series is a single time series.
While time series data are assumed to be univariate and finite length, stream data is presented in both univariate and multivariate form while having unknown length.
The taxonomy provided by \cite{gupta_outlier_2014} is convoluted and incomplete for a number of reasons. 
Given the different category names, it is implied that data streams are not time series.
Furthermore, it is implied that time-series data cannot be multivariate.
In contrast, the taxonomy provided in Section \ref{sec:taxonomy} covers all of the cases mentioned by \cite{gupta_outlier_2014} in a more streamlined and intuitive way.}

In addition to that, \cite{aggarwal_time_2017} discusses time-series anomaly detection in an online setting. 
The two anomaly types identified in time series by the author are referred to as contextual and collective anomalies. 
\lc{They define contextual anomalies as time steps that significantly deviate from their temporal neighbours, i.e.\ \hyperref[subsec:anomaly_types]{\emph{point anomalies}}. 
Collective anomalies are defined as an unusually shaped collection of contiguous time steps, which encompass both \hyperref[subsec:anomaly_types]{\emph{sub-sequence and whole-sequence anomalies}} and hence make no distinction between them.}

\cite{mason_online_2019} published a survey on what is considered traditional methods, i.e.\ not based on deep learning (DL). 
These methods are categorised into four domains: statistics, time-series analysis, pattern mining and machine learning. 
They proceed to further classify each of the approaches into parametric and non-parametric. 
In addition, they comment on the online applicability of the methods and provide an overview of machine learning platforms.

DL has gained more traction in the last few years, due to its scalability and good performance when presented with larger amounts of data and the reduced need for feature engineering  \cite{chalapathy_deep_2019}.

The first paper that compares traditional methods with DL-driven anomaly detection for online applications is published by \cite{munir_comparative_2019} where the authors survey several methods from both paradigms and provide a quantitative evaluation and comparison between the presented models for two different publicly available data sets. 
They conclude that despite not being the fastest approach, DL provides the best performance for the chosen performance metrics.

\cite{blazquez-garcia_review_2021} published a survey encompassing both DL and classical methods. 
They provide a novel taxonomy which deviates slightly from the point, collective and contextual anomaly convention, first suggested by \cite{chandola_anomaly_2009}. 
\lc{Here the authors classify anomalies into three categories: point outliers, sub-sequence outliers and outlier time series, renamed to \hyperref[subsec:anomaly_types]{\emph{point, sub-sequence and whole-sequence anomalies}} for the sake of consistency in this work.}
Furthermore, they discuss the idea of applying univariate techniques to multivariate data, which is not discussed in previous publications. 

\lc{\cite{lindemann_survey_2021} create a survey specialising in LSTM-based models for anomaly detection. 
They classify the approaches by input data properties, model architecture and evaluation procedure and provide an outlook on graph-based and transfer learning methods.
A taxonomy on time-series anomaly detection is also provided, though it closely follows the one by \cite{chandola_anomaly_2009}.
The paper does not delve into a discussion on data sets and mentions evaluation metrics very briefly without deeper analysis.}

\lc{Similar to \cite{munir_comparative_2019}, \cite{ngo_bibinbe_survey_2022} provides a comparative study between mostly traditional approaches and one deep learning-based approach in the context of streaming data. 
Of the data sets used for evaluation, only one is multivariate and the evaluation metrics are kept fairly simple, while also adding tolerance to predicted labels if they are close enough to ground-truth labels, making the results essentially incomparable for other studies unless the same tolerance is used.}

\lc{While extensive, the analysis provided by \cite{sgueglia_systematic_2022} focuses solely on approaches for anomaly detection in Internet of Things. A short overview of the public data sets is provided, though only key facts are presented without a critical perspective. Additionally, there is no mention of the proposed evaluation metrics in the literature. Like \cite{lindemann_survey_2021}, they provide anomaly detection taxonomy, though not novel, as it is first proposed by \cite{chandola_anomaly_2009}.}

\cite{li_deep_2023} identify three types of anomalies: anomalous time points, anomalous time intervals and anomalous time series, analogous to \hyperref[subsec:anomaly_types]{\emph{point, sub-sequence and whole-sequence anomalies}}. 
They then provide an overview of deep learning methods for time-series anomaly detection sorted by type of anomaly. 
Certain properties such as architecture, threshold type and presence of feature extraction are commented on, as well as what metrics are chosen for evaluation in each of the publications\lc{, though no deeper analysis on them is undertaken.}
Lastly, they dedicate a chapter to data sets suitable for benchmarking\lc{, however, like with metrics, no further analysis is presented.}

Several issues are identified in the current survey landscape. 
\lc{In the past, \cite{chandola_anomaly_2012} and \cite{gupta_outlier_2014} presented strict and convoluted definitions of \hyperref[subsec:continuous_discrete_sequence]{\emph{discrete-sequence and continuous-sequence anomaly detection}}.
Furthermore, previous work fails to distinguish between online training and online inference, which are characteristics important for real-world applications.
While two of the surveys provide a brief overview of data sets used for benchmarking anomaly detection algorithms, they fail to point out the associated problems.
In addition to that, there is no survey providing a complete overview of the evaluation metrics proposed, never mind a deeper analysis comparing them with each other.}
Hence, this survey aims to provide the following contributions:
\begin{itemize}
	\item Novel taxonomy in the area of online anomaly detection, making a distinction between online training and online inference
    \item \lc{Homogenised taxonomy of continuous- and discrete-sequence anomaly detection problems}
	\item \lc{Extensive overview and analysis of the most popular benchmark data sets used}
    \item \lc{Extensive overview and analysis of proposed evaluation metrics}
	\item Updated overview on the state-of-the-art methodology used for model-based online anomaly detection
\end{itemize}

\section{Benchmarking} \label{sec:benchmark}
Benchmarking is an important part of anomaly detection research, as it enables objectively measuring progress in the field.
To benchmark any given approach against others, it has to be applied to the same data set and use the same evaluation metrics.
This section gives a detailed overview of the most commonly used public data sets and their respective weaknesses, as well as of the metrics used to evaluate anomaly detection approaches and proposed improvements in the literature.

\subsection{Data Sets}\label{subsec:data_sets}
Anomaly detection literature uses a variety of time-series data sets over the years.
Publicly available data sets play an important role in enabling researchers to benchmark their methods against the state of the art.
To inform the reader about the publicly available data sets, the key information is summarised in Table \ref{tab:data}.
\begin{table}[h!]
\centering
\begin{tabular}{lcccccc}
\hline
Name & $d_\mathcal{D}$ & $\#O$ & $N_n$  & $N_a$ & UTS        & $\%A$  \\ \hline\hline
HRSS & 18              & 1   & 2      & 2     &            & 24\% \\
TEP  & 56              & 2   & 546    & 142   &            & 15\% \\
WADI & 86              & 5   & 1      & 1     &            & 6\%  \\
SWaT & 45              & 9   & 1      & 1     &            & 12\% \\
SMAP & 25              & 10  & 54     & 54    &            & 13\% \\
MSL  & 55              & 10  & 27     & 27    &            & 11\% \\
CNC  & 47              & 1   & 8      & 10    &            & 53\% \\
SMD  & 38              & 6   & 28     & 28    & \checkmark & 4\%  \\
MSDS & 10              & 1   & 1      & 1     & \checkmark & 72\% \\
PSM  & 10              & 2   & 1      & 1     & \checkmark & 28\% \\ \hline
\end{tabular}
\caption{Key information about the most popular data sets summarised. The key is as follows, \lc{
$d_\mathcal{D}$: dimensionality of the data set, i.e.\ number of features, $\#O$: number of occurrences in publications considered in this survey, $N_n$: the number of nominal sequences, $N_a$: number of anomalous sequences, UTS: unlabelled training subset, $\%A$ the number of anomalous time steps in relation to all test time steps.}}
\label{tab:data}
\end{table}

Unfortunately, most popular data sets are unsuitable for time-series anomaly detection, as concluded by \cite{wu_current_2021}.
They make a case against many publicly available data sets by pointing out four main flaws, at least one of which can be found in most data sets.
These flaws include:
\begin{itemize}
    \item triviality
    \item unrealistic anomaly density
    \item uncertain labels
    \item run-to-failure bias
\end{itemize}
They define triviality as being able to be solved with a very limited amount of code, hence weakening the case for the need for complex parameter-heavy models.
Then they point out the unrealistic anomaly density, where the assumption that anomalies are very rare events does not hold.
The next issue \cite{wu_current_2021} discuss is the apparent mislabelled ground truth.
In some cases, regions of similar behaviour are sometimes labelled anomalous, sometimes labelled as nominal, which can skew results.
Lastly, they argue that some data sets have a run-to-failure bias, i.e.\ anomalies appear at the end of sequences and hence the anomaly detection performance can be improved by "guessing" that the last time steps are anomalous.
Perhaps the word failure is inaccurate in this context, as an anomaly is not necessarily a failure but simply a deviation from nominal behaviour.
An alternative and more general term could be \emph{nominal-to-anomaly} bias.
It could be argued that it is nominal-to-anomaly bias is intrinsic in anomaly detection as it is more likely for a system to behave nominally and then fail than the other way around, although this depends on the application.
The aforementioned issues are only pointed out for a subset of the public data sets considered by \cite{wu_current_2021}, however, their criticism can be extended to all other public data sets, which is discussed in the following.

The High Rack Storage System (HRSS) data set \cite{dataset_hrss} represents the behaviour of an unoptimised (\textit{standard}) and \textit{optimised} factory at the Ostwestfalen-Lippe University of Applied Sciences.
One issue present is that at a few points, nominal time steps are sandwiched between anomalous time steps, raising suspicion of possible mislabelling.
\begin{figure}[h!]
    \centering
    \includegraphics[width=\textwidth]{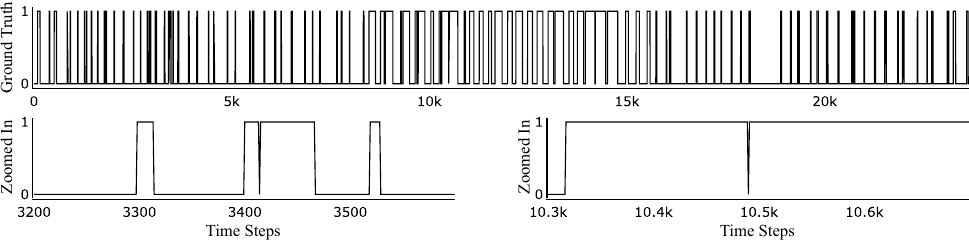}
    \caption{Ground-truth label vector from the HRSS data set plotted, as well as two subplots zoomed into different time windows. A nominal time step between two anomalous time steps could represent mislabelling.}
    \label{fig:hrss}
\end{figure}
It features an unrealistic anomaly density, as around 24\% of the test time steps are labelled as anomalous, as shown in Figure \ref{fig:hrss}.
\lc{Lastly, two nominal and two anomalous sequences are provided, but no specification of training and testing subsets, therefore publications using this data set may not be comparable.}

The Kaspersky Lab provides a data set from the process industry.
Using an industrial process model, they propose the Tennessee-Eastman Process data set (TEP) \cite{dataset_tep}\lc{, for which a labelled nominal training subset is provided.}
As specified by \cite{filonov_rnn-based_2017}, the \textit{DANGER} warnings are considered the anomalies.
The TEP data set has a few anomalous sequences that suffer from unrealistic anomaly density, where a single nominal time step is sandwiched between two anomalous sub-sequences.
In addition to that, the type 21 anomalies can easily be detected by looking at the feature 8.

The Singapore University of Technology and Design published two data sets: the Secure Water Treatment (SWaT) \cite{dataset_swat} and the Water Distribution (WADI) \cite{dataset_wadi} data sets.
Both consist of test bed data under nominal conditions and under various attack scenarios, however, WADI is composed of higher dimensional data, though no training/testing subset splits are provided.
SWaT and WADI are not explicitly named by \cite{wu_current_2021}, but still suffer from the same flaws.
For example, the SWaT data set suffers from a variety of issues.
First and foremost, it contains half a dozen features that are constant throughout the train and test sets, making them redundant for modelling and hence are not considered in Table \ref{tab:data}.
In addition to that, it features a very high anomaly density, in fact, the anomaly between time steps 227828 and 262727 accounts for around 65\% of the total anomalous time steps.
\begin{figure}[h!]
    \centering
    \includegraphics[width=\textwidth]{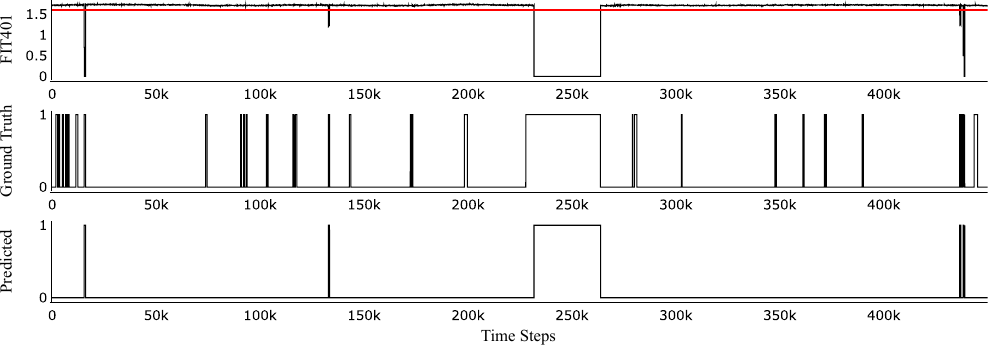}
    \caption{FIT401 channel, as well as ground-truth and predicted label vectors from the SWaT data set plotted. Setting a threshold of 1.6 (red line in the first subplot) yields a sample-based precision of 0.99, a sample-based recall of 0.63 and a sample-based $F_1$ score of 0.76.}
    \label{fig:swat}
\end{figure}
This anomaly is considered trivial as it can easily be detected by just looking at a single feature, as shown in Figure \ref{fig:swat}.
Of the 127 features in WADI many need to be filtered out.
Like SWaT, WADI suffers from redundant features, 33 in total, as well as features with missing values, 4 in total, and features with NaN values, another 4 in total, bringing the total feature count to 86.
Of the four approaches found in the literature that use this data set, all use a different feature count, making this incomparable.

NASA published two commonly used data sets: the Soil Moisture Active Passive (SMAP) satellite data set and the Mars Science Laboratory (MSL) rover data set \cite{dataset_smap, dataset_msl}.
\lc{Both are split into training and testing subsets, though the training subset consists of only nominal samples.}
In the supporting material of \cite{wu_current_2021}, they indicate that both data sets suffer from triviality and unrealistic anomaly density.
Apart from the afore-mentioned criticism, it should be noted that, while technically multivariate in the sense that the sequences in either data set do have multiple channels, all channels other than the first one (the telemetry channel) are binary one-hot encoded command channels with no dynamic behaviour.
\cite{hundman_detecting_2018} only predict the one telemetry channel in their work, which may be acceptable for predictive models but not applicable to other model families, like reconstructive models.

The University of Michigan published the CNC Mill Tool Wear data set \cite{dataset_cnc}, which consists of 18 multivariate time series.
The main issue with this data set is that labels only exist for an entire sequence (worn/unworn) and from the 18 sequences only 8 are nominal (unworn).
Given that some of the eight sequences may need to be used as training data, the data set is unbalanced, but towards the anomaly class.
Because the anomaly class is more common in the test set, the precision metric will be inflated, as false positives will naturally be small.
Lastly, the CNC data set features a very high anomaly density of 53\%, hence, it is not recommended for anomaly detection.

Another multivariate time-series data set is the Server Machine Data set (SMD) \cite{dataset_smd} which features 38 channels \lc{ and unlabelled training subsets for each machine.}
Like SMAP and MSL, SMD also suffers from triviality and unrealistic anomaly density.
\cite{wu_current_2021} illustrate the triviality by detecting anomalies using simple statistical metrics, such as the moving standard deviation.

The Multi-Source Distributed System (MSDS) data set \cite{dataset_msds} is specifically generated for the development of anomaly detection, root cause analysis, and remediation methods, according to the authors.
One stand-out characteristic is that it features per-channel anomaly labels for the testing subset, however, it has a big shortcoming: it has a completely unrealistic anomaly density of 72\%.
\begin{figure}[h!]
    \centering
    \includegraphics[width=\textwidth]{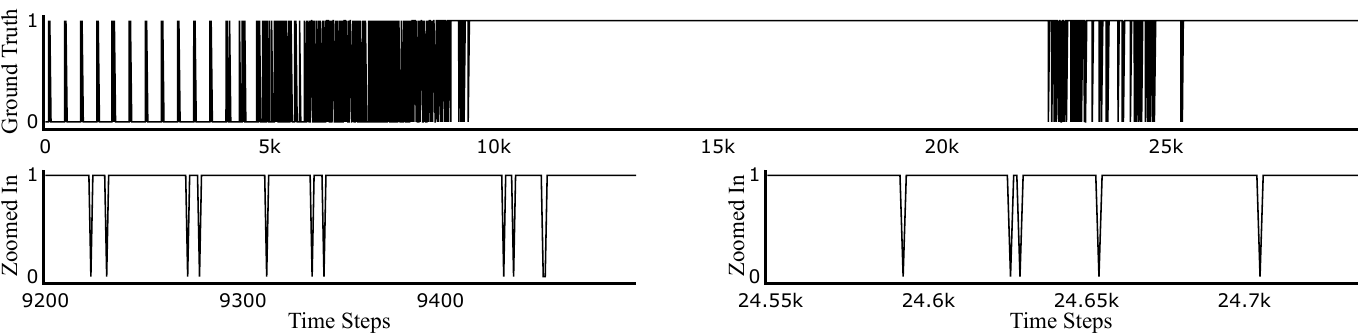}
    \caption{Ground-truth label vector from the MSDS data set plotted, as well as two subplots zoomed into different time windows. 72\% of the time steps are labelled as anomalous, which is far too high given the rare nature of anomalies. \lc{In addition to that, there are a number of instances of uncertain labelling.}}  
    \label{fig:msds}
\end{figure}
Its longest sub-sequence anomaly (starting at around time step 10000 in Figure \ref{fig:msds}) accounts for 44\% of the test sequence and 61\% of the anomalous time steps within that sequence, making it inappropriate for anomaly detection research. \lc{Furthermore, there are several instances where a ground-truth nominal time step is sandwiched between ground-truth anomalous time steps, indicating dubious ground-truth labels.}

EBay published a data set called the Pooled Server Metrics (PSM) data set \cite{dataset_psm}.
It consists of several channels depicting server machine metrics like CPU utilisation and memory used, similar to SMD.
\begin{figure}[h!]
    \centering
    \includegraphics[width=\textwidth]{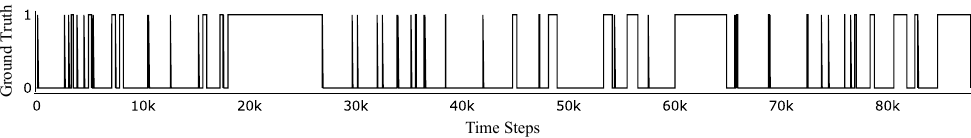}
    \caption{Ground-truth label vector from the PSM data set plotted. 28\% of the time steps within are labelled as anomalous, which is not the worst rate of the data sets surveyed, but is still too large.}
    \label{fig:psm}
\end{figure}
In most channels, many of the maximum and minimum values correspond to anomalies (Figure \ref{fig:psm}), hence a case could be made for the triviality of the data set.
Furthermore, it features a fairly high anomaly density of 28\%.

For clarity, the issues present in each data set are highlighted in Table \ref{tab:pros_cons_datasets}.
\begin{table}[h!]
\centering
\begin{tabular}{lcccc}\hline
Name & TA & UAD & UL & SI \\ \hline\hline
HRSS &   & \checkmark    & \checkmark &  \\
TEP  & \checkmark  & \checkmark    &  &  \\
WADI &   &     &   & \checkmark \\
SWaT & \checkmark  & \checkmark     &  & \checkmark  \\
SMAP &   & \checkmark    &  &  \\
MSL  &   & \checkmark    &  &  \\
CNC  &   & \checkmark    &  &  \\
SMD  & \checkmark  & \checkmark    &  &  \\
MSDS &   & \checkmark    & \checkmark &  \\
PSM  & \checkmark  & \checkmark    &   & \\ \hline
\end{tabular}
\caption{\lc{A table summarising the problems identified in each of the data sets presented in this subsection. Key is as follows, TA: trivial anomalies, UAD: unrealistic anomaly density, UL: uncertain labels, SI: structural inconsistencies.}} \label{tab:pros_cons_datasets}
\end{table}

As a solution to the benchmarking problems outlined \lc{in this subsection}, \cite{wu_current_2021} propose using a collection of 250 sequences, the Hexagon ML/UCR Time Series Anomaly Detection data set.
While addressing the identified flaws, this collection exclusively features univariate sequences, hence it cannot be recommended by the authors of this survey as a realistic representation of complex real-world anomaly detection tasks occurring in, for example, the manufacturing industry.
Therefore, there is still a clear need for a high-quality multivariate time-series data set \lc{ with pre-defined training and testing subsets, as well as clearly described training subset in terms of contamination levels to enable a just and reliable comparison between approaches.}

\subsection{Metrics} \label{subsec:metrics}
Evaluation metrics allow different anomaly detection performance aspects to be quantified and compared.
These metrics can generally be separated into two different classes: calibrated and uncalibrated.
In order to provide a consistent notation throughout the following subsections, the following notation system is introduced.
Reconsider a testing subset $\mathcal{D}^{\text{test}}$ in data set $\mathcal{D}$ containing $N$ sequences that can be nominal (without anomalies), entirely anomalous or partially anomalous, such that $\mathcal{D}^{\text{test}} = [\mathcal{S}_1, ..., \mathcal{S}_n, ..., \mathcal{S}_{N}]$, where $\mathcal{S}_n \in \mathbb{R}^{T_{n} \times d_\mathcal{D}}$.
Here, $T_n$ is the number of time steps in sequence $\mathcal{S}_n$, which varies depending on $\mathcal{S}_n$ and $d_\mathcal{D}$ the number of features, i.e.\ the dimensionality of $\mathcal{S}_n$.
The ground-truth label set $L^{\text{gt}}$ corresponding to $\Gamma$ contains a binary label vector for each sequence, such that $L^{\text{gt}} = [\boldsymbol{l}_1^{\text{gt}}, ..., \boldsymbol{l}_n^{\text{gt}}, ..., \boldsymbol{l}_N^{\text{gt}}]$, where $\boldsymbol{l}_n^{\text{gt}} \in \mathbb{B}^{T_{n}}$.
Anomalous time steps are assigned a $1$ and nominal time steps a $0$.
Therefore, for a given nominal sequence $\sum_{t=0}^{T_n} l_{n,t}^{\text{gt}} = 0$, for an entirely anomalous sequence $\sum_{t=0}^{T_n} l_{n,t}^{\text{gt}} = T_n$ and for a partially anomalous sequence $0 < \sum_{t=0}^{T_n} l_{n,t}^{\text{gt}} < T_n$.
As a counterpart to the ground-truth label set $L^{\text{gt}}$, there is the corresponding predicted label set $L^{\text{p}}$, such that $L^{\text{p}} = [\boldsymbol{l}_1^{\text{p}}, ..., \boldsymbol{l}_n^{\text{p}}, ..., \boldsymbol{l}_N^{\text{p}}]$, where $\boldsymbol{l}_n^{\text{p}} \in \mathbb{B}^{T_{n}}$.

\subsubsection{Calibrated metrics}\label{subsubsec:calibrated_metrics}
Similar to binary classification, anomaly detection generally segregates two classes using a threshold, except that one class is far rarer than the other, hence can be seen as imbalanced classification.
\lc{This threshold is often applied to some sort of anomaly score.
Generally, the anomaly score is some sort of error metric as a function of time, most of the time it is closely related to the loss function.}
Likewise, correct predictions can be labelled as true positives and true negatives and incorrect ones into false positives and false negatives, where negative refers to a nominal case and positive to an anomalous case.
From here on, the number of true positives is represented by $N_\text{tp} \in \mathbb{W}$, the number of true negatives by $N_\text{tn} \in \mathbb{W}$, the number of false positives by $N_\text{fp} \in \mathbb{W}$ and the number of false negatives by $N_\text{fn} \in \mathbb{W}$, all of which are whole numbers.
For now, label vector temporality is ignored and each time step is considered individually as a sample, therefore all anomalous time steps are treated as individual point anomalies, i.e.\ $S=1$.
Depending on the match between the predicted label vector $\boldsymbol{l}_n^{\text{p}}$ and ground-truth label vector $\boldsymbol{l}_n^{\text{gt}}$, the number of true positives, true negatives, false negatives and true positives can be obtained; see Equation \ref{eq:tp_fp_tn_fn}.
\begin{equation}
    \begin{split}
        N_\text{tp} = \sum_{n=1}^{N} \boldsymbol{l}_n^{\text{p}} \cdot \boldsymbol{l}_n^{\text{gt}} \quad \quad N_\text{tn} = \sum_{n=1}^{N} (\boldsymbol{o}_n-\boldsymbol{l}_n^{\text{p}}) \cdot (\boldsymbol{o}_n-\boldsymbol{l}_n^{\text{gt}})\\
        N_\text{fn} = \sum_{n=1}^{N} (\boldsymbol{o}_n-\boldsymbol{l}_n^{\text{p}}) \cdot \boldsymbol{l}_n^{\text{gt}} \quad \quad N_\text{fp} = \sum_{n=1}^{N} \boldsymbol{l}_n^{\text{p}} \cdot (\boldsymbol{o}_n-\boldsymbol{l}_n^{\text{gt}})
    \end{split}
    \label{eq:tp_fp_tn_fn}
\end{equation}
where $\boldsymbol{o}_n$ is a vector of ones the same size as $\boldsymbol{l}_n^{\text{gt}}$ and $\boldsymbol{l}_n^{\text{p}}$ such that $\boldsymbol{o}_n \in \mathbb{B}^{T_{n}}$.

Precision $P$ is a metric that shows the number of true positives in relation to the total number of flags, i.e.\ how many of the flags are correct.
Precision $P$ can be calculated using Equation \ref{eq:precision}. As is evident, a smaller number of false positives leads to a higher precision.
\begin{equation}
    P = \frac{N_\text{tp}}{N_\text{tp}+N_\text{fp}}
    \label{eq:precision}
\end{equation}

Recall $R$ is often presented in combination with precision, as it measures the number of false negatives in relation to the total anomaly count, i.e.\ how many of the anomalies are detected.
Recall can be calculated using Equation \ref{eq:recall}. As is evident, a smaller number of false negatives leads to a higher recall.
\begin{equation}
    R = \frac{N_\text{tp}}{N_\text{tp}+N_\text{fn}}
    \label{eq:recall}
\end{equation}

Precision and recall can be summarised into a single metric, the $\mathrm{F}_\beta$ score.
It takes the weighted harmonic mean of the two metrics; see Equation \ref{eq:f_beta}.
\begin{equation}
    \mathrm{F_\beta} = \frac{(1+\beta^2)\cdot N_\text{tp}}{(1+\beta^2)\cdot N_\text{tp} + \beta^2\cdot N_\text{fn}+ N_\text{fp}} = (1+\beta^2)\cdot\frac{P\cdot R}{(\beta^2\cdot P)+R}
    \label{eq:f_beta}
\end{equation}
Typically, precision and recall are equally weighted, hence $\beta=1$ and we speak of the $F_1$ score; see Equation \ref{eq:f1}.
\begin{equation}
    \mathrm{F_1} = \frac{N_\text{tp}}{N_\text{tp} + \frac{1}{2}\cdot(N_\text{fn} + N_\text{fp})} = 2\cdot\frac{P\cdot R}{P+R}
    \label{eq:f1}
\end{equation}
The metrics \lc{in Equations \ref{eq:tp_fp_tn_fn}-\ref{eq:f1}} are referred to as calibrated metrics since they represent anomaly detection performance at a specific threshold/sensitivity.

One metric that is often used in binary classification is the accuracy $\Phi$, which represents the total number of correct classifications in relation to all samples, as shown in Equation \ref{eq:acc}.
\begin{equation}
    \Phi = \frac{N_\text{tp} + N_\text{tn}}{N_\text{tp} + N_\text{fn} + N_\text{fp} + N_\text{tn}}
    \label{eq:acc}
\end{equation}
However, accuracy is unsuitable for anomaly detection, given the imbalanced nature of the problem. For example, in a case with 90 nominal and 10 anomalous samples and an anomaly detector that classifies all nominal samples correctly, but none of the anomalous ones would have an accuracy figure of 90\%, despite it detecting $0$ anomalies.

One alternative to regular accuracy is balanced accuracy $\Phi_B$, which calculates the accuracy in both classes separately and then takes the average, as in Equation \ref{eq:bacc}.
While more suitable for anomaly detection, this performance metric has seen no mention in any of the literature surveyed in this work.
\begin{equation}
    \Phi_B = \frac{1}{2}\left(\frac{N_\text{tp}}{N_\text{tp} + N_\text{fn}}+\frac{N_\text{tn}}{N_\text{tn} + N_\text{fp}}\right)
    \label{eq:bacc}
\end{equation}

As previously mentioned, these metrics are generally used in imbalanced classification and are sample-based, i.e.\ each time step is considered individually.

The first proposal for metrics apt to evaluate sub-sequence anomalies is given by \cite{xu_unsupervised_2018}.
They apply the metrics such that a ground-truth anomalous sub-sequence is considered a true positive if it overlaps with any predicted anomalous time step.
False positives and false negatives are obtained as previously.
This method is generally referred to as point-adjustment.
Despite its prominence in the literature, point-adjusted metrics come with a number of problems.
Firstly, in a case where only the last anomalous time step is flagged, a significant delay in the detection is present which can be undesired.
Furthermore, it makes anomaly detection trivial.
In the case of a single long sub-sequence anomaly within a time series or a whole-sequence anomaly, a single time step within would lead to a true positive, despite all other time steps not being classified correctly.
\begin{figure}[h!]
    \centering
    \includegraphics[width=\textwidth]{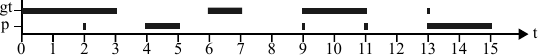}
    \caption{Plot of arbitrary ground-truth and predicted label vectors $\boldsymbol{l}_n^{\text{gt}}$ and $\boldsymbol{l}_n^{\text{p}}$, respectively.}
    \label{fig:tp_fp_fn}
\end{figure}
To underline the effect this has on metrics, consider Figure \ref{fig:tp_fp_fn}.
Without point-adjust, the figure shows $N_\text{tp}=4$, $N_\text{fp}=4$, $N_\text{tn}=2$ and $N_\text{fn}=6$, and hence a precision of 0.50, a recall of 0.4 and therefore an $\mathrm{F_1}$ score of 0.44.
With point-adjustment, however, the figure shows $N_\text{tp}=8$, $N_\text{fp}=4$, $N_\text{tn}=2$ and $N_\text{fn}=2$, and hence a precision of 0.66 and a recall of 0.80 and therefore an $\mathrm{F_1}$ score of 0.72.
Clearly, point-adjusted metrics are not well suited for sub-sequence anomaly detection, as they are too forgiving and ignore delays in detection.
In fact, \cite{doshi_reward_2022} show that sampling a uniform distribution parameterised with an alarm probability can outperform most state-of-the-art models when point-adjusted metrics are used.

\cite{tatbul_precision_2018} propose so-called range-based recall $R_\mathrm{r}$ and precision $P_\mathrm{r}$.
For the range-based recall $R_\mathrm{r}$ calculation, four aspects are relevant to anomalous sub-sequences within a time series.
The first is existence, which is analogous to the point-adjust way of marking a true positive.
The corresponding existence score $\epsilon$ for each ground-truth anomaly is $1$ if any predicted positive time steps overlap with it and $0$ otherwise.
In the context of Figure \ref{fig:tp_fp_fn} the existence score is $1$ for the first, third and fourth ground-truth anomaly and $0$ for the second one.
Then, there is the position, i.e.\ what part of the predicted anomaly overlaps with the true anomaly.
For this aspect, this method requires an application-related bias $\delta$.
There may be use cases where correctly flagging certain sections within the true anomaly and therefore the front, back, middle, or all time steps should be emphasised, though it is unclear why the back end of an anomaly should be rewarded.
The next aspect is size, i.e.\ how close the size of the predicted anomaly is to the true anomaly, denoted by the overlap size $\omega$.
In Figure \ref{fig:tp_fp_fn} the overlap size is low for the first ground-truth anomalous sub-sequence and $1$ for the third and fourth one.
For the second ground-truth anomaly, the overlap size is undefined, as there is no predicted time step that overlaps with it.
The overlap size is also a function of the aforementioned application-related bias $\delta$.
Lastly, there is cardinality.
One example of a cardinality score $\gamma$ would be one that indicates how many predicted positive sub-sequences overlap with each ground-truth positive sub-sequence.
This is especially useful to punish approaches that flag several positive sub-sequences that fit inside a single ground-truth positive sub-sequence rather than a single continuous one.
In Figure \ref{fig:tp_fp_fn} the cardinality score for the first and fourth ground-truth anomalous sub-sequence would be higher than for the third one.
Again, for the second one, this would be undefined due to the lack of corresponding predicted positive time steps.
The cardinality factor is used to scale the overlap size figure, yielding the overlap score which is combined with the existence score using a pre-defined weight $\alpha$ to yield the range-based recall for each ground-truth anomaly $i$, as shown as high-level in Equation \ref{eq:r_r}.
\begin{equation}
    R_\mathrm{r, i} = \alpha\cdot\epsilon + (1-\alpha)\cdot\gamma\cdot\omega
    \label{eq:r_r}
\end{equation}
For the range-based precision $P_\mathrm{r}$ the existence score does not apply and hence only the overlap, position, and cardinality scores are considered for each predicted anomaly.
The aforementioned scores are then applied inversely to the above, i.e.\ for each predicted anomaly $j$, rather than for each ground-truth anomaly $i$.
Finally, the individual range-based recalls and precisions are averaged for all ground-truth anomalies and all predicted anomalies to obtain the final range-based recall $R_\mathrm{r}$ and precision $P_\mathrm{r}$, respectively.
While this way of evaluating anomaly detection performance is fairly robust, it assumes that in a test time series, there is always a predicted anomaly, which may not always be the case.
In a case where this does not happen, the method runs into a numerical error.
Furthermore, it requires the setting of bias $\gamma$, as well $\alpha$, which sets the balance between existence and size, position, and cardinality.

\cite{hwang_time-series_2019} propose an extension to the point-adjust metrics, called the time-aware precision $P_\mathrm{{ta}}$ and time-aware recall $R_\mathrm{{ta}}$.
For their metrics, \cite{hwang_time-series_2019} assume that each positive ground-truth has an ambiguous tail of length $\delta$, where the time steps in $T<t<T+\delta$ are considered anomalous too.
Both time-aware precision $P_\mathrm{{ta}}$ and time-aware recall $R_\mathrm{{ta}}$ consist of two components, the detection score and the portion score.
The detection score components of the time-aware precision and recall are $P^\mathrm{d}_\mathrm{ta}$ and $R^\mathrm{d}_\mathrm{ta}$, respectively. 
The way true positives are counted is similar to the point-adjust score, except that it requires the predicted anomaly to overlap with the ground truth by a set threshold $\theta$.
The portion score components $P^\mathrm{p}_\mathrm{ta}$ and $R^\mathrm{p}_\mathrm{ta}$ are defined as the average overlap between ground truth and predicted anomalies.
The final time-aware precision $P_\mathrm{{ta}}$ and recall $R_\mathrm{{ta}}$ is obtained using the weighted sum of the detection and portion scores, respectively, parameterised by weight $\alpha$.
Like range-based recall $R_\mathrm{r}$ and precision $P_\mathrm{r}$, time-aware precision $P_\mathrm{{ta}}$ and recall $R_\mathrm{{ta}}$, require a manually defined parameter $\alpha$.

To specifically address the delay in detection, \cite{doshi_reward_2022} propose a metric called the average detection delay $\bar D$, which is the mean delay for an approach to label a ground-truth anomaly as such over all ground-truth anomalies.
For example, the detection delay $D$ for the first ground-truth anomaly in Figure \ref{fig:tp_fp_fn} would be two samples and zero for the third and fourth ground-truth anomalies.
To cap the delay time of detections, a maximum tolerable delay $\tau_{max}$ is used if the detection takes longer than $\tau_{max}$, therefore the worst $\bar D$ score possible is $\bar D=\tau_{max}$.
The average detection delay $\bar D$ can be normalised using $\tau_{max}$, in which case the worst value is $\tilde D=\bar D/\tau_{max}=1$.
It is undesirable to flag time steps too early, hence \cite{doshi_reward_2022} propose the sequence alarm precision $P_\text{SA}$, where a false positive is recorded for early flags.
It is calculated the same way regular precision is, except that a false positive is defined as an early flag, therefore $P_\text{SA}$ is the number of early flags relative to all flags.
Like the two proposed metrics discussed previously, the manual setting of $\tau_{max}$ is required.
Furthermore, it is unclear what the delay should be if there is no predicted anomaly in a given test sequence.
Also, \cite{doshi_reward_2022} do not propose a calibrated metric that combines $\tilde D$ and $P_\text{SA}$, like $F_1$ combines precision $P$ and recall $R$.

In an attempt to create a set of interpretable and parameter-free metrics, \cite{huet_local_2022} propose the following.
First, a local affiliation between the ground-truth label vector $\boldsymbol{l}_n^{\text{gt}}$ and the predicted label vector $\boldsymbol{l}_n^{\text{p}}$ is established.
This is done by splitting test sequences into chunks, each with exactly one ground-truth anomaly.
For the precision in each chunk, they calculate the average temporal distance $d_P$ between any false positive time steps and the closest ground-truth positive time step.
When this average temporal distance is zero, it means that no time steps within the chunk are false positives.
In the case of the recall, this is flipped, so that the average temporal distances $d_R$ between any false negative time steps and the closest ground-truth time step are calculated.
When this average temporal distance is zero, it means that no time steps within the chunk are false negatives.
Then, the average directed distances are averaged over all chunks, resulting in $P_\text{td}$ and $R_\text{td}$, since the temporal distances are interpreted as precision and recall.
To provide context on the anomaly detection performance, the absolute temporal distances can be replaced with sample-based precision and recall probabilities $p_P$ and $p_R$, respectively.
These are normalised against random predictions, hence a value of above $0.5$ in either means that they are better than random predictions.
Like the metrics proposed previously, \cite{huet_local_2022} do not address cases where no predicted anomalies are present in a chunk.

\cite{wagner_timesead_2023} propose several improvements to the metrics introduced by \cite{tatbul_precision_2018}.
The first adjustment involves the weight parameter $\alpha$.
They point out that the existence score is superfluous as the other scores already imply the existence, i.e.\ overlap of ground truth and predicted positive sub-sequence, hence $\alpha$ is set to $0$.
Secondly, they suggest a new cardinality score that is consistent across all types of bias, however, they point out the importance of the constant bias due to its neutrality.
Lastly, \cite{wagner_timesead_2023} change the way the individual precision scores are combined.
\cite{tatbul_precision_2018} simply take the mean of all precisions, but rather than dividing the sum of precisions by the number of predictions, \cite{wagner_timesead_2023} use a weighted sum of precisions, which depend on the length of the respective prediction.
These improvements lead to a parameter-free evolution of the range-based precision and recall, at least when constant bias is considered the default, however, the new metrics still suffer from the same assumption that for every ground-truth anomaly, there is at least one predicted anomaly.

To avoid ambiguity on how multiple sub-sequence anomalies within a sequence should be detected and counted, \cite{wu_current_2021} suggest framing the anomaly detection problem such that each test sequence has a single anomaly that is either detected or not detected.
Data sets with a single test time series containing multiple anomalies would require splicing to create multiple sub-sequences, each with a single anomaly.
\begin{table}[h!]
\centering
\begin{tabular}{lccccccc}\hline
Publications                        & PA     & SSA & WSA & CSAD          & DSAD           & PF & NAPP \\ \hline\hline
Xu et al. \cite{xu_unsupervised_2018}   & \checkmark & \checkmark   & \checkmark  & \checkmark          &                      & \checkmark     & \checkmark             \\
Tatbul et al. \cite{tatbul_precision_2018}  & \checkmark & \checkmark   & \checkmark  & \checkmark          &                      &                &                        \\
Hwang et al. \cite{hwang_time-series_2019} & \checkmark & \checkmark   & \checkmark  & \checkmark          &                      &                &                        \\
Doshi et al. \cite{doshi_reward_2022}      & \checkmark & \checkmark   &             & \checkmark          &                      &                &                        \\
Huet et al. \cite{huet_local_2022}        & \checkmark & \checkmark   & \checkmark  & \checkmark          & \checkmark           & \checkmark     &                        \\
Wagner et al. \cite{wagner_timesead_2023}   & \checkmark & \checkmark   & \checkmark  & \checkmark          &                      & \checkmark     &                        \\\hline    
\end{tabular}
\caption{Table summarising the key weaknesses and strengths of the metrics proposed in Subsection \ref{subsubsec:calibrated_metrics}. The key is as follows, PA: point anomaly, SSA: sub-sequence anomaly, WSA: whole-sequence anomaly, CSAD: continuous-sequence anomaly detection, DSAD: discrete-sequence anomaly detection, PF: parameter-free, NAPP: no assumption of positive prediction}
\label{tab:metrics}
\end{table}

Table \ref{tab:metrics} summarises the key distinguishing factors of the metrics introduced before.
As is evident, almost all methods are compatible with all types of anomalies, with the sequence alarm precision $P_\text{SA}$ by \cite{doshi_reward_2022} being the exception.
Recall that $P_\text{SA}$ represents the number of early flags relative to all flags.
In the case of time-series anomalies, this cannot be applied, since it is impossible to flag a time step early.
All metrics are introduced in the context of continuous-sequence anomaly detection and therefore it is nowhere specified how they can be extended to discrete-sequence anomaly detection.
Despite not mentioning discrete-sequence anomaly detection, \cite{huet_local_2022} split the test time series into chunks like \cite{wu_current_2021} suggest, making the method, in theory, compatible with discrete-sequence anomaly detection.
Only \cite{xu_unsupervised_2018, huet_local_2022, wagner_timesead_2023} offer non-parametric solutions, which are preferred for benchmarking, as it removes tunable elements that can alter results and hinder fair comparison.
Lastly, all metrics but the point-adjusted ones \cite{xu_unsupervised_2018} require an existing prediction for each ground-truth anomaly, else they return numerical errors.
This is especially relevant in discrete-sequence anomaly detection, where a whole-sequence anomaly may be entirely labelled as nominal.
Despite the extensive work done in the area, there is still a clear need for a set of metrics that can handle all anomalies and anomaly detection types and do not require parameters or predicted anomalies.

\subsubsection{Uncalibrated metrics} \label{subsubsec:uncalibrated_metrics}
Metrics that do not require a threshold are called uncalibrated metrics.
Rather than representing anomaly detection performance at a certain point, they show the potential of general anomaly detection performance, whereas calibrated metrics show the performance when the approach is applied.
The most prominent metric is the area under the precision-recall curve $A_{\text{PR}}$.
The precision-recall curve is a function of precision with respect to recall and hence, in the continuous realm, $A_{\text{PR}}$ can be calculated using Equation \ref{eq:apr}.
\begin{equation}
    A_\text{PR}^\text{cont}=\int_0^1 P \,dR 
    \label{eq:apr}
\end{equation}

In practice, however, the PR curve is discrete and therefore the area under it can be obtained using a variety of discrete integration methods, such as the midpoint rule, the trapezoidal rule or Simpson's rule.
In the case of the trapezoidal rule, the discrete area under the precision-recall curve is obtained using Equation \ref{eq:disc_apr}.
\begin{equation}
    A_\text{PR}^\text{disc}=\frac{1}{2}\sum^{K}_{k=1}(P_{k-1}+P_{k}) \cdot \Delta R_k
    \label{eq:disc_apr}
\end{equation}
\begin{figure}[h!]
    \centering
    \includegraphics[width=0.5\textwidth]{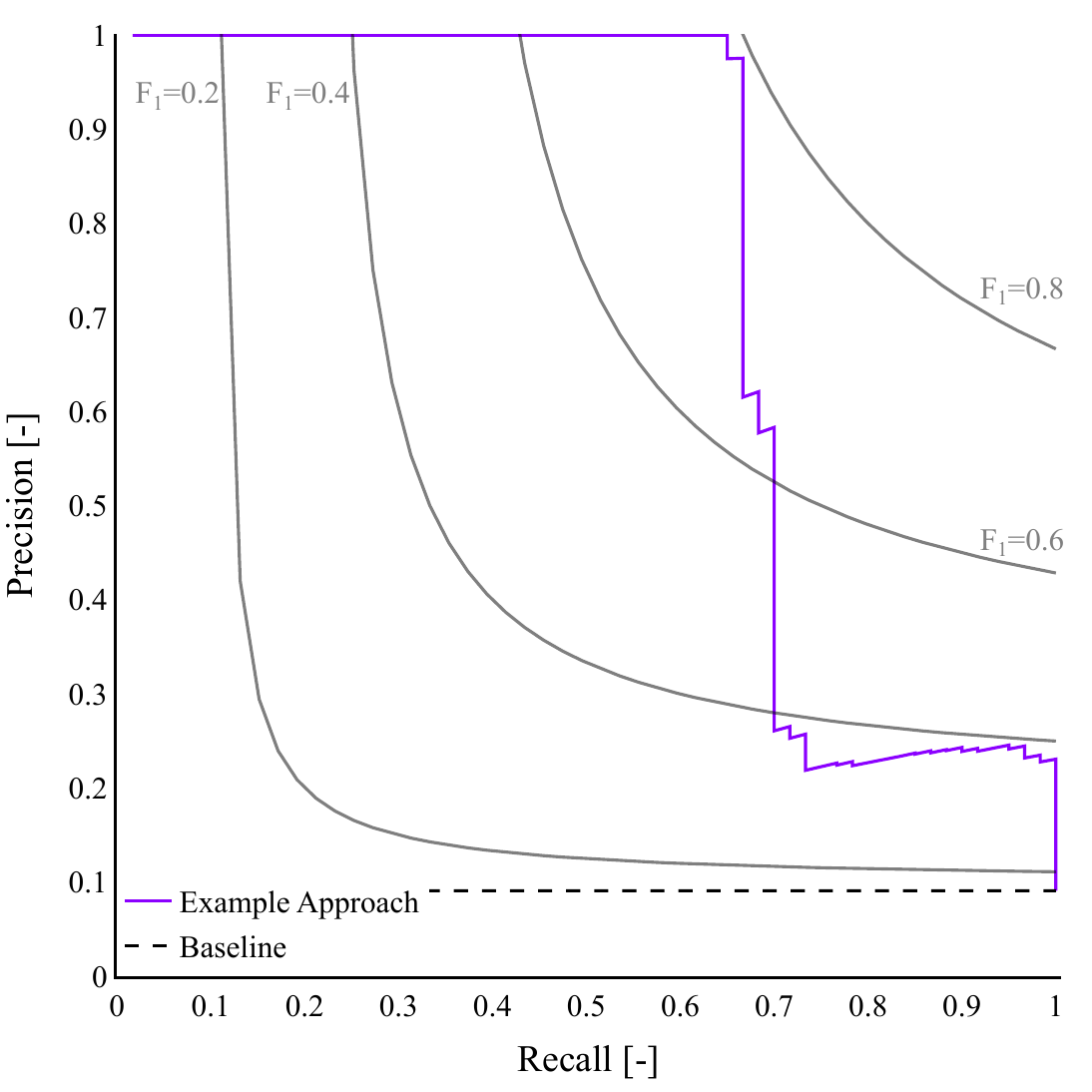}
    \caption{Example of the precision-recall curve for an unspecified approach. When calibrated, the approach would be placed on a single point along this curve. The closer it is to the $P=R=1$ point, the higher the $F_1$ score.}
    \label{fig:apr}
\end{figure}
An example of a precision-recall curve is shown in Figure \ref{fig:apr}.

The average precision $\bar P$ can be seen as an alternative method of reducing a precision-recall curve to a single scalar.
It is calculated as the weighted mean precision over all values along the precision-recall curve, as shown in Equation \ref{eq:ap}.
As far as the authors are aware, only \cite{audibert_deep_2022} use this metric for evaluating anomaly detection approaches.
\begin{equation}
    \bar P=\sum^{K}_{k=1}P_{k} \cdot \Delta R_k
    \label{eq:ap}
\end{equation}
In essence, this is just a discrete way of integrating the precision-recall curve and finding the area under it. The average precision metric therefore simply denotes the $A_\text{PR}$ when right-endpoint integration is used, rather than the mid-point integration rule.

In addition to the $A_{\text{PR}}$, there is the area under the receiver operating characteristic curve $A_{\text{ROC}}$, which is mainly used in the context of balanced binary classification.
This curve is a function of the true-positive rate with respect to the false-positive rate.
Similar to $\Phi$, $A_{\text{ROC}}$ is unsuitable for imbalanced classification and therefore anomaly detection, as in the same example the false positive rate is disproportionally influenced by the majority class.

\cite{doshi_reward_2022} suggest combining both the normalised average detection delay $\tilde D$ and sequence alarm precision $P_\text{SA}$ by obtaining the area under the $P_\text{SA}$-$\tilde D$ curve $A_{P_\text{SA}\tilde D}$, similar to the $A_\text{PR}$.

\cite{paparrizos_volume_2022} propose a set of metrics which, like \cite{hwang_time-series_2019}  account for label uncertainty.
In their work, they propose transforming the binary label vector into a continuous one.
First, they split the test time series into chunks \cite{huet_local_2022}, each with a single ground-truth anomaly within.
Then, \cite{paparrizos_volume_2022} extend each ground-truth sub-sequence by adding an ambiguous tail to the end of the sub-sequence, like \cite{hwang_time-series_2019}, and an ambiguous head to the beginning, both of length $\psi$.
The values inside the ambiguous head and tail are not binary, however, they are assigned a gradient.
When two sub-sequence anomalies are less than $\psi$ apart, then the tail and the head, respectively, are superimposed.
Also, the length of the ambiguous head and tail $\psi$ can be set to a variety of values: half the period of the sequence, the average length of the sub-sequence anomalies or a manually defined value.
The metrics \lc{in this subsection} are obtained the same way as shown in Equation \ref{eq:tp_fp_tn_fn}, i.e.\ sampled-based, however, as is evident, these are no longer whole numbers but instead rational numbers, i.e.\ $N_\text{tp} \in \mathbb{Q}$, $N_\text{fp} \in \mathbb{Q}$, $N_\text{tn} \in \mathbb{Q}$ and $N_\text{fn} \in \mathbb{Q}$.
While this reduces interpretability, precision and recall can still be calculated as shown in Equations \ref{eq:precision} and \ref{eq:recall}.
Furthermore, the sum of these metrics still equates to the number of samples in the ground-truth label vector $\boldsymbol{l}_n^{\text{gt}}$.
Based on this underlying evaluation system, \cite{paparrizos_volume_2022} propose two uncalibrated metrics: the range-based area under the precision-recall curve $A_\text{RPR}$ and the volume under the precision-recall surface $V_\text{PR}$.
Calculating the $A_\text{RPR}$ is trivial, as it is obtained the same way $A_{\text{PR}}$ is found, i.e.\ using Equation \ref{eq:disc_apr}.
$V_\text{PR}$ is also similar to $A_\text{RPR}$.
In essence, it is an extension that takes a variable ambiguous head and tail length $\psi$ into account, turning a 2D curve into a 3D surface.
The area under it is therefore found by integrating across two axes, rather than just one.
The discrete integration is done using Equation \ref{eq:disc_vpr}.
\begin{equation}
    V_\text{PR}^\text{disc}=\frac{1}{4}\sum^{\Psi}_{\psi=1}\sum^{K}_{k=1}(P_{\psi, k-1}+P_{\psi, k}) \cdot \Delta R_{\psi, k}
    \label{eq:disc_vpr}
\end{equation}
Given that the period of the sequence and the average length of the sub-sequence anomalies are not known a priori, $\psi$ will default to a manually defined value, turning this set of metrics into essentially parametric.
In addition to that $N_\text{tp}$, $N_\text{fp}$, $N_\text{tn}$ and $N_\text{fn}$ are not as interpretable as their classic counterparts.

Like with the calibrated metrics, there is no uncalibrated metric that represents anomaly detection performance for a range of thresholds, while also being compatible with all anomaly and anomaly detection types, non-parametric and non-assuming of anomaly predictions.

\section{Predictive Models}
\label{sec:predictive_modelling}
Predictive models are models that are trained to predict a time step given a finite window $\mathbf{W} \in \mathbb{R}^{w\times d_\mathcal{D}}$ of past values. Consider predictive model $\mathcal{M}$, which in most publications is trained such that for time-series window $\mathbf{W}$ it takes all but the last time step of $\mathbf{W}$ and predicts the last time step of $\mathbf{W}$, as shown in Equation \ref{eq:pred}.
\begin{equation}
    \hat{\mathbf{W}}_w = \mathcal{M}(\mathbf{W}_{0:w-1})
    \label{eq:pred}
\end{equation}
Such predictive models can be used for anomaly detection by comparing their predictions with the observed/measured value using a variety of techniques to obtain an error metric.
The key assumption is that nominal, i.e.\ anomaly-free, sequences are predicted with a small error, whereas anomalies will lead to a large error between prediction and observed values.
Predictive models tend to be the simplest and on average the oldest model family discussed in this survey.
An overview of all the prediction-based modelling publications is presented in Table \ref{tab:summary_pred}.
The keys in the table are chosen such that the reader can quickly recognise what type of approach is taken by the respective author.
\begin{table}[h!]
\centering
\begin{tabular}{lcccc}
\hline
Publications                       & OT         & OI         & DwL        & EM                    \\ \hline\hline
Yen et al. \cite{yen_causalconvlstm_2019}    & \checkmark & \checkmark &            & $F_1$                 \\
Que et al. \cite{que_real-time_2019}         &            & \checkmark &            & CM                    \\
Munir et al. \cite{munir_deepant_2019}         &            & \checkmark & \checkmark & $F_1$, $A_\text{ROC}$ \\
Malhotra et al. \cite{malhotra_long_2015}         &            &            &            & $F_{0.1}$             \\
Chauhan and Vig \cite{chauhan_anomaly_2015}       &            &            &            & $F_{0.1}$             \\
Filonov et al. \cite{filonov_multivariate_2016}  &            &            &            & $F_1$                 \\
Filonov et al. \cite{filonov_rnn-based_2017}     &            &            & \checkmark & -                     \\
Thill et al. \cite{thill_anomaly_2019}         &            &            &            & $F_1$                 \\
Hundman et al. \cite{hundman_detecting_2018}     &            &            & \checkmark & $F_{0.5}$             \\
Tambuwal and Neagu \cite{tambuwal_deep_2021}         &            &            &            & $F_1$                 \\
Sakuma and Matsutani \cite{sakuma_area-efficient_2021} &            &            & \checkmark & $A_\text{ROC}$        \\ 
Pan et al. \cite{pan_duma_2023}              &            &            &            & $F_1$                 \\
Xia et al. \cite{xia_coupled_2024}           &            &            &            & $F_1$                 \\ \hline
\end{tabular}
\caption{Summary of all publications surveyed in the area of prediction-based time-series anomaly detection. The key is as follows, \lc{OT: online training, OI: online inference, DwL: detection without labels, EM: evaluation metrics, CM: confusion matrix.}}
\label{tab:summary_pred}
\end{table}

\subsection{Online Training and Online Inference}
\label{subsec:ononpred}
\cite{yen_causalconvlstm_2019} present an anomaly detection algorithm for computer logging records, that convolutes the input log key sequence using four parallel convolutional neural networks (CNN) \cite{fukushima_neocognitron_1980, lecun_gradient-based_1998} layers with varying kernel sizes.
The approach is trained on nominal data only, hence we speak of semi-supervised anomaly detection.
The resulting feature maps, i.e.\ the vector representation of the CNN output, are then concatenated and fed into a long short-term memory (LSTM) \cite{hochreiter_long_1997, gers_learning_2000} layer which then predicts a probability distribution across all log keys.
The most likely log keys are considered nominal, where the minimum likelihood at which a key is considered nominal is the threshold.
In their work, they refer to the threshold as a hyperparameter, implying that it is obtained empirically with labels.
\cite{yen_causalconvlstm_2019} experiment with different online training techniques, including regularly retraining independently of anomaly detection results and retraining if the false-positive rate exceeds a given value.

\subsection{Offline Training and Online Inference}
\cite{que_real-time_2019} bring forward a method to detect anomalies during flight testing of commercial aircraft.
It utilises the encoder component of a trained autoencoder to reduce the dimensionality of a multi-variate signal.
Following that, a pair of LSTM layers predict the next time step based on the latent vector output by the transferred encoder.
The error resulting from these predictions is fit to a Gaussian distribution, then allowing for a likelihood to be estimated, hence anomalies can be detected if the likelihood exceeds a given confidence interval.
This confidence interval is an assumption made, hence this approach is considered unsupervised anomaly detection.

\cite{munir_deepant_2019} propose a stacked CNN network, which predicts the following time step based on a given history window.
The anomaly score is based on the Euclidean norm (also known as the L2 norm) of the difference between the observed value and the prediction, and therefore needs to be evaluated against a threshold.
\cite{munir_deepant_2019} mention a parametric and a non-parametric way to obtain a threshold, though in the results, it is not specified which of the thresholds is used.

\subsection{Offline Training and Offline Inference but Online Capable}
The majority of existing work makes use of, sometimes stacked, recurrent neural networks (RNN) to predict future time steps, mostly trained on nominal data, except for \cite{tambuwal_deep_2021, sakuma_area-efficient_2021}.
These predictions are then held up against the observed value, from which an anomaly score is calculated.
The main difference is the error metrics used as the anomaly score.

\cite{malhotra_long_2015} and \cite{chauhan_anomaly_2015} use the prediction error resulting from a validation set to fit a Gaussian distribution using maximum likelihood estimation.
During testing, they calculate the probability that the obtained error belongs to the fitted distribution and then compare it with a threshold found using labelled data.

\cite{filonov_multivariate_2016} and \cite{filonov_rnn-based_2017} both use the mean-squared error (MSE) between observation and prediction as the anomaly score, although \cite{filonov_rnn-based_2017} smooths the anomaly score using an exponential weighted moving average (EWMA), with a smoothing factor.
\cite{filonov_multivariate_2016} \lc{treat the threshold as a tunable hyperparameter}, whereas \cite{filonov_rnn-based_2017} set the threshold to the $99.9^{th}$ percentile of the smoothed error resulting from the training set.
\lc{On its own, the threshold choice by \cite{filonov_rnn-based_2017} does not require labels to be set, though it is not justified why specifically the $99.9^{th}$ percentile is chosen.}

\cite{thill_anomaly_2019} apply a variety of post-processing steps to the LSTM-predictor output, \lc{where the model is fit on nominal data only.}
First, they remove outliers from the absolute prediction error by discarding the top and bottom 3\% prediction errors to allow for a good fitting of a Gaussian distribution.
Using the resulting prediction error, the squared Mahalanobis distance is calculated.
To further refine the detection process, \cite{thill_anomaly_2019} apply a correction procedure which, for each time step, chooses the prediction within a horizon which results in the lowest absolute error.
The threshold is chosen in an $F_1$ score-maximising manner based on the labelled test data\lc{, though a threshold-free way to detect anomalies based on Rosner's outlier test \cite{rosner_percentage_1983} is proposed too.}

\cite{hundman_detecting_2018} propose using a dynamic error threshold on the smoothed prediction error to detect anomalies in spacecraft sensor data.
This is done by computing a dynamic threshold for every time step using a custom function that takes in the mean and standard deviations of the EWMA-smoothed absolute reconstruction error window, as well as an adjustable parameter that adjusts the sensitivity.

Rather than just predicting the next time step, \cite{tambuwal_deep_2021} present an implementation of a stacked LSTM network that outputs quantiles for that prediction.
Modelling this uncertainty is facilitated using a custom loss function that is based on the quantile values.
Using the predicted quantiles, a confidence interval can be calculated using the difference between the upper and lower quantiles.
Hence, this confidence interval can be used to find anomalies, which would be indicated by a large confidence interval.
\lc{The threshold is chosen such that it achieves a desired trade-off between precision and recall using labelled validation data.}

Another rather unique approach is taken by \cite{sakuma_area-efficient_2021}, where a sparse recurrent neural network, known as an echo state network, is used to make next-time-step predictions.
The loss corresponding to the time step of the prediction is then turned into the anomaly score.
Hotelling’s T\textsuperscript{2} test is used to determine whether a given anomaly score is anomalous or not.

\lc{\cite{pan_duma_2023} proposes a prediction model consisting of a 1D CNN layer, a self-attention and a dense layer in that order. 
The approach masks input sequences before the forward pass and then applies max-masking in the self-attention block. 
Masking is a regularisation process where a specified amount of random sub-sequences in the data are discarded to force the model to interpolate between known sub-sequences. 
Max-masking, specifically, involves retaining the top $k$\% values in the attention score matrix and hence requires the $k$ parameter to be set.
The output of the model is the sum of the output of the dense layer and a so-called highway, which is a linear transformation of the input window.}

\lc{\cite{xia_coupled_2024} proposes an encoder-decoder model which features an additional decoder for prediction. 
Given the different learning tasks for each of the decoders, the loss function consists of a term minimising the prediction error and another minimising the reconstruction error. 
Curiously, the anomaly score is calculated using the prediction error only, hence why this approach is classified as a predictive model in this survey.
The threshold used to segment the anomaly score into nominal and anomalous time steps is obtained by $F_1$ score-maximising grid search.}

\section{Reconstructive Models}
\label{sec:reconstructive_modelling}
Reconstructive modelling is usually based on some variation of an autoencoder.
Autoencoders consist of an encoder $\mathcal{M}_\text{Enc}$ and decoder $\mathcal{M}_\text{Dec}$, which generally mirror each other in architecture.
The encoder compresses a given input time series window $\mathbf{W} \in \mathbb{R}^{w\times d_\mathcal{D}}$ into a latent vector $\mathbf{z} \in \mathbb{R}^{u\times d_\mathbf{z}}$, i.e.\ a reduced representation of the input space.
Depending on the approach, this latent representation $\mathbf{z}$ may or may not contain temporality, i.e.\ $u=1$ or $u>1$, respectively. The latent vector is then expanded again to reconstruct the input $\mathbf{W}$ through the decoder, as shown in Equation \ref{eq:rec}.
\begin{eqnarray}
    \begin{split}
        \mathbf{z} &= \mathcal{M}_\text{Enc}(\mathbf{W}) \\
        \hat{\mathbf{W}} &= \mathcal{M}_\text{Dec}(\mathbf{z})
    \end{split}
    \label{eq:rec}
\end{eqnarray}

Reconstructive models rely on the assumption that the reconstruction error will be large when faced with anomalous data.
An overview of the reconstruction-based modelling approaches discussed in this survey is presented in Table \ref{tab:summary_rec}.
\begin{table}[h!]
\centering
\begin{tabular}{lcccc}
\hline
Publications                                 & OT & OI         & DwL        & EM                            \\ \hline\hline
Gugulothu et al. \cite{gugulothu_sparse_2018}                &    & \checkmark &            & $A_\text{ROC}$                \\
Hsieh et al. \cite{hsieh_unsupervised_2019}              &    & \checkmark &            & $F_1$                         \\
Audibert et al. \cite{audibert_usad_2020}                   &    & \checkmark &            & $F_1$                         \\
Bayram et al. \cite{bayram_real_2021}                     &    & \checkmark &            & $F_1$, $A_\text{ROC}$         \\
Malhotra et al. \cite{malhotra_lstm-based_2016}             &    &            &            & $F_{0.1}$, $F_{0.05}$         \\
Hmayouni et al. \cite{homayouni_autocorrelation-based_2020} &    &            &            & $F_1$                         \\
Lindemann et al. \cite{lindemann_anomaly_2020}               &    &            &            & -                             \\
Nguyen et al. \cite{nguyen_forecasting_2021}              &    &            & \checkmark & $\Phi$, $F_1$                 \\
Naito et al. \cite{naito_anomaly_2021}                   &    &            &            & $F_1$                         \\
Kieu et al. \cite{kieu_outlier_2018}                    &    &            &            & $F_1$                         \\
Kieu et al. \cite{kieu_outlier_2019}                    &    &            &            & $A_\text{PR}$, $A_\text{ROC}$ \\
Tayeh et al. \cite{tayeh_attention-based_2022}           &    &            & \checkmark & $F_1$                         \\
Zhang et al. \cite{zhang_deep_2019}                      &    &            &            & $F_1$                         \\
Chadha et al. \cite{chadha_deep_2021}                     &    &            &            & $F_1$, CM                     \\
Thill et al. \cite{thill_temporal_2021}                  &    &            & \checkmark & $F_1$                         \\
Gong et al. \cite{gong_prediction-augmented_2021}       &    &            &            & $F_1$                         \\
Zhang et al. \cite{zhang_reconstruct_2021}               &    &            &            & $A_\text{ROC}$                \\ \hline
\end{tabular}
\caption{Summary of all publications surveyed in the area of reconstruction-based time-series anomaly detection. The key is as follows,  \lc{OT: online training, OI: online inference, DwL: detection without labels, EM: evaluation metrics, CM: confusion matrix.}}
\label{tab:summary_rec}
\end{table}

\subsection{Online Training and Online Inference}
\label{subsec:ononrec}
Approaches based on autoencoders that are trained and infer in an online fashion do not exist, to the best of the author's knowledge.
Technically such methods are possible, presumably they would work very similarly to the approach in Subsection \ref{subsec:ononpred}.
It is difficult to estimate how well this hypothetical algorithm would work, though it would be an open research direction to pursue.

\subsection{Offline Training and Online Inference}
The first approach that involved offline training of an autoencoder that is capable of online inference is proposed by \cite{gugulothu_sparse_2018}\lc{, which is trained on nominal data only}.
It bears a lot of similarity to the work by \cite{malhotra_lstm-based_2016} and adds a feature reduction layer between the input layer and the encoder, which acts as a regulariser in capturing dependencies between channels.
The encoder and decoder are both based on LSTM layers, whose output, i.e.\ reconstruction, is then evaluated using the Mahalanobis distance as the anomaly score, which is held up against a threshold found through $\mathrm{F_1}$ score maximisation to detect anomalies.

\cite{hsieh_unsupervised_2019} use an LSTM-based autoencoder \lc{trained on nominal data} to detect anomalies in smart manufacturing via the mean squared error.
They applied transfer learning to improve the algorithm performance significantly.
This is done by pre-training the model on data from different points in the manufacturing process.

Inspired by the training process of GANs, \cite{audibert_usad_2020} propose using two dense autoencoders $\mathcal{M}_1$ and $\mathcal{M}_2$ which share the same encoder.
Training occurs in two phases, with the first one focusing on reconstruction error minimisation of the input window for both decoders.
The second phase involves pitting the two autoencoders against one another.
$\mathcal{M}_1$ attempts to minimise the error between the input and $\mathcal{M}_2$' reconstruction of $\mathcal{M}_1$' reconstruction while $\mathcal{M}_2$ attempts to maximise the reconstruction error $\mathcal{M}_2$' reconstruction of $\mathcal{M}_1$' reconstruction
The reconstruction error used throughout this paper is the mean-squared error.
This leads to $\mathcal{M}_1$ being explicitly trained to create reconstructions that, when input into $\mathcal{M}_2$, lead to outputs very similar to the input.
$\mathcal{M}_2$ on the other hand, is trained to be able to distinguish between $\mathcal{M}_1$' reconstruction and the input window.
The anomaly score for an input window during inference is then a weighted sum between $\mathcal{M}_1$' reconstruction error and $\mathcal{M}_2$' reconstruction error of $\mathcal{M}_1$' reconstruction.
\lc{\cite{audibert_usad_2020} investigate the anomaly detection performance when the training set is contaminated with different amounts of anomalous data.}
It, too, has to be mentioned that for very dynamic test sets this approach may not work well, as it cannot model temporality.

\cite{bayram_real_2021} compare a convolutional autoencoder with a novel autoencoder architecture to detect acoustic anomalies in industrial processes.
It combines convolutional layers with convolutional LSTM layers \cite{shi_convolutional_2015}, which rely on the convolution operation rather than matrix multiplication.
The Euclidean distance is then computed from the reconstruction.
The threshold is taken as the value that yields the maximum difference between the true-positive rate and the true-negative rate.

\subsection{Offline Training and Offline Inference but Online Capable}
\cite{malhotra_lstm-based_2016} propose using an LSTM autoencoder which reconstructs input windows in reverse temporal order and use the Mahalanobis distance as the anomaly score.
A threshold is obtained by maximising a modified version of the $\mathrm{F}$ score, with weighted precision over recall, using a small amount of labelled data.

\cite{homayouni_autocorrelation-based_2020} introduce an LSTM autoencoder-based approach which uses autocorrelation to systematically find a suitable window length.
The time-series data is windowed according to the result of an autocorrelation analysis, which finds the length of the dynamics within the signal.
For each feature in a time series, the autocorrelation at each lag is calculated.
The autoencoder is capable of ingesting both unlabelled and labelled data to improve the detection of anomalies.
This is done by adding an extra label column as one of the features to be reconstructed.
The label value varies depending on whether the time step is faulty, suspicious, unknown, or valid.
In addition to that, they utilise a decision tree model in an attempt to provide explainability to the anomalies, by isolating the channel that most likely causes the anomaly.

\cite{lindemann_anomaly_2020} employ an LSTM autoencoder with discrete wavelet transforms at each end to better intake data of different frequencies.
After training \lc{on nominal data}, the encoder is removed and the decoder is used as an inverse model for the metamodel, a Nonlinear Autoregressive with External Input network.
The difference between the metamodel input and the decoder output is then used to detect anomalies.
\lc{Quantitative results, such as precision, recall or $F_1$ score are not provided.}

\cite{nguyen_forecasting_2021} use an LSTM autoencoder paired with a one-class support vector machine to detect anomalies.
It works by separating nominal and anomalous data points in the absolute error vector resulting from the reconstruction using a hyperplane\lc{, hence a threshold is not required.}

\cite{naito_anomaly_2021} use two autoencoders, one to reconstruct the time series, and another one to reconstruct the reconstruction error from the previous autoencoder, which is then added to the reconstructed time series, in theory yielding a superior reconstruction.
\lc{The anomaly score used is the L2 norm with segmented by a threshold set such that the $F_1$ score is maximised.}

\cite{kieu_outlier_2018} compare a 2D CNN autoencoder with an LSTM autoencoder for anomaly detection in enriched time series.
These time series are augmented with derived and statistical features, hence enlarging their feature space compared to the raw time series.
In addition to that, the autoencoder reconstruction is enhanced by embedding one-hot encoded contextual information in it.

\cite{kieu_outlier_2019} then propose using an ensemble of sparse recurrent autoencoders.
The autoencoders differ from each other as they use different sparseness weight vectors.
Furthermore, two types of ensembles are proposed: an independent framework, where the autoencoders are run independently and a shared framework, where the latent vector is shared between autoencoders.
Anomalies are then detected using the median value from the Euclidean distance vector of all autoencoders and a manually defined threshold.

\cite{zhang_deep_2019} attempt to detect anomalies by creating a convolutional autoencoder that works with correlation matrices, here called signature matrices, rather than raw time series.
The autoencoder is further enhanced by skip connections that connect layers in the encoder with their decoder counterparts in order to allow the model to better deal with long sequences.
The skip connections process the information flowing through them, by running it through a ConvLSTM layer paired with an unnamed custom attention layer.
The algorithm can provide a root cause analysis by labelling the channels associated with the three worst reconstructed correlations in a given matrix.

\cite{tayeh_attention-based_2022} propose a similar an autoencoder structure similar to \cite{zhang_deep_2019}, though without the skip connections.
Here, a Bahdanau-style attention mechanism \cite{bahdanau_neural_2015} is added to the model to maintain performance with increasing length of input sequences.
The approach can dynamically adjust its anomaly detection threshold, similar to \cite{hundman_detecting_2018}, which is used to label time steps, in this case, correlation matrices, as anomalies.
Like what is proposed by \cite{zhang_deep_2019}, this algorithm can provide a degree of explainability to its outputs, though by using a threshold to detect suspicious channels within the correlation matrix.

\cite{chadha_deep_2021} expand on the idea of using a convolutional autoencoder for anomaly detection by adding a clustering element to the latent space.
The latent space is split into a discriminative and reconstructive component which is used for clustering and reconstruction, respectively.
The clustering element regularises the latent space by adding a clustering loss term to the total loss function.
In addition to that, the authors propose a semi-supervised variation of the model where the trained decoder is replaced by dense layers that work as a classifier.

\cite{thill_temporal_2021} suggest a temporal convolutional network (TCN) autoencoder with a series of improvements.
For example, the authors suggest reversing the dilation rate ordering at the decoder, to better adapt it to the lower resolution representation of the time series in the latent space.
Inspired by the ResNet \cite{he_deep_2016} and DenseNet \cite{huang_densely_2017} architectures, they implement skip connections to prevent an overreliance of the model on the high dilation rate representation.
Furthermore, \cite{thill_temporal_2021} also propose using the reduced representations between hidden layers to detect anomalies, as they represent different frequency scales due to the variable dilation rate.
In addition to that, they use map reduction layers after each hidden layer to reduce the dimensionality of the convolution and therefore the number of tunable parameters.
Lastly, they propose running the anomaly score through a high-pass filter set to cut off frequencies lower than 1Hz to remove drifts from the signal.
The anomaly score used is the Mahalanobis distance.

\cite{gong_prediction-augmented_2021} propose a different way to implement skip connections in a convolutional autoencoder.
Rather than moving the information through a ConvLSTM and an attention layer at the skip connection, like \cite{zhang_deep_2019}, the information is processed by a multi-layer perceptron layer followed by a Bahdanau-style attention layer.
Another innovation is the use of a prediction module at the bottleneck of the autoencoder to enhance the latent space mapping during training.
The prediction module consists of a multi-layer perceptron which predicts the following input window from the latent space.
The loss function is customised to include both reconstruction and prediction loss, requiring a weight parameter that varies the ratio between either loss.

\cite{zhang_reconstruct_2021} develop an adversarial approach to train an anomaly detection network.
Here the model consists of a convolutional autoencoder and a deconvolutional classifier, which, in a way, compete against each other during training.
Nominal samples are corrupted using a custom algorithm parameterised through a corruption level to represent anomalies and are then input into the autoencoder along with their uncorrupted counterpart.
Once the latent vector is computed for either sample, the latent vector error is found and minimised.
The goal of this is to create as plausible a reconstruction from the latent vector as possible, regardless of the input sample.
This reconstruction is then used as a class to train the classifier, which is used to tell the nominal, uncorrupted sample apart from the reconstruction.
The loss function used for the autoencoder is composed of the reconstruction loss and the latent vector error multiplied by a pre-defined weight.

\section{Generative Models}
\label{sec:generative_modelling}
Generative models can be further segmented into two most common model types: VAEs \cite{kingma_auto-encoding_2014} and GANs \cite{goodfellow_generative_2014}.
The former bears some similarity to traditional autoencoders, other than the fact that the low-dimensionality representation is not mapped to a vector $\mathbf{z}$ but rather a distribution $(\boldsymbol{\mu}_\mathbf{z}, \boldsymbol{\sigma}_\mathbf{z}) \in \mathbb{R}^{u \times d_\mathbf{z}}$ from which a latent vector $\tilde{\mathbf{z}} \in \mathbb{R}^{u \times d_\mathbf{z}}$ is sampled; see Equation \ref{eq:gen}. 
\begin{eqnarray}
    \begin{split}
        (\mu_\mathbf{z}, \sigma_\mathbf{z}) &= \mathcal{M}_\text{Enc}(\mathbf{W})\\ 
        \mathbf{z}_\text{samp} &\sim  (\boldsymbol{\mu}_\mathbf{z}, \boldsymbol{\sigma}_\mathbf{z})\\
        \hat{\mathbf{W}} &= \mathcal{M}_\text{Dec}(\mathbf{z}_\text{samp})
    \end{split}
    \label{eq:gen}
\end{eqnarray}
Again, depending on the publication, the latent vector may or may not contain temporality.
In some cases, the output of the decoder is not deterministic, but instead a distribution $(\boldsymbol{\mu}_\mathbf{W}, \boldsymbol{\sigma}_\mathbf{W})$ that approximates the input $\mathbf{W}$.
In the generative modelling literature, the encoder of an autoencoder is also referred to as the recognition model and the decoder as the generative model.
GANs, on the other hand, work by training a generator $\mathcal{M}_\text{Gen}$ network to generate a plausible time series window $\hat{\mathbf{W}} \in \mathbb{R}^{w \times d_\mathcal{D}}$ sampled from a normal distribution $\mathcal{N}(0, 1)$.
A second network $\mathcal{M}_\text{Disc}$ called the discriminator then tries to classify the generated input window $\hat{\mathbf{W}} \in \mathbb{R}^{w \times d_\mathcal{D}}$ and the training example $\mathbf{W}$ correctly; see Equation \ref{eq:gan}.
\begin{eqnarray}
    \begin{split}
        \mathbf{z}_\text{samp} &\sim \mathcal{N}(0, 1)\\
        \hat{\mathbf{W}} &= \mathcal{M}_\text{Gen}(\mathbf{z}_\text{samp})\\
        \mathbf{l} &= \mathcal{M}_\text{Disc}(\mathbf{W}) \\
        \mathbf{l} &= \mathcal{M}_\text{Disc}(\hat{\mathbf{W}})
    \end{split}
    \label{eq:gan}
\end{eqnarray}

As training progresses, the generator becomes better at generating samples, hence discriminating between the generated input and the real data becomes harder and harder.
However, the discriminator also becomes better at telling real samples apart from generated samples.
Hence, training can be thought of as a two-player min-max game \cite{goodfellow_generative_2014}.
An overview of all the generation-based modelling publications is presented in Table \ref{tab:summary_gen}.
\begin{table}[h!]
\centering
\begin{tabular}{lcccc}
\hline
Publications                      & OT         & OI         &  DwL       & EM                                   \\ \hline\hline
Suh et al. \cite{suh_echo-state_2016}       & \checkmark & \checkmark &            & $F_1$                                \\
Park et al. \cite{park_multimodal_2018}      &            & \checkmark & \checkmark & $A_\text{ROC}$                       \\
Su et al. \cite{su_robust_2019}            &            & \checkmark & \checkmark & $F_1$                                \\
Zhang et al. \cite{zhang_federated_2021}      &            & \checkmark & \checkmark & $F_1$                                \\
Li et al. \cite{li_mad-gan_2019}           &            & \checkmark &            & $F_1$                                \\
Fan et al. \cite{fan_luad_2023}             &            & \checkmark & \checkmark & $F_1$                                \\
Zhang et al. \cite{zhang_acvae_2024}          &            & \checkmark & \checkmark & $F_1$                                \\
Guo et al. \cite{guo_multidimensional_2018} &            &            &            & $\Phi$, $F_1$, $A_\text{ROC}$        \\
von Schleinitz et al. \cite{von_schleinitz_vasp_2021}  &            &            &            & $A_\text{ROC}$                       \\
Li et al. \cite{li_anomaly_2021}           &            &            &            & $F_1$, $A_\text{ROC}$, $A_\text{PR}$ \\
Choi et al. \cite{choi_multivariate_2022}    &            &            &            & $F_1$                                \\
Zhou et al. \cite{zhou_beatgan_2019}         &            &            &            & $A_\text{ROC}$, $\bar P$             \\
Choi et al. \cite{choi_gan-based_2020}       &            &            & \checkmark & -                                    \\
Geiger et al. \cite{geiger_tadgan_2020}        &            &            &            & $F_1$                                \\
Zhu et al. \cite{zhu_novel_2019}            &            &            &            & $\Phi$, $F_1$, $A_\text{ROC}$        \\
Sun et al. \cite{sun_time_2019}             &            &            & \checkmark & -                                    \\ \hline
\end{tabular}
\caption{Summary of all publications surveyed in the area of generative-based time-series anomaly detection. The key is as follows,  \lc{OT: online training, OI: online inference, DwL: detection without labels, EM: evaluation metrics, CM: confusion matrix.}}
\label{tab:summary_gen}
\end{table}

\subsection{Online Training and Online Inference}
\cite{suh_echo-state_2016} propose a VAE based on an echo state network in an effort to adapt it for anomaly detection in time-series data.
This model is trained in an online manner, i.e.\ the parameters are updated as new data is input.
The anomaly score is given by the negative log-likelihood of an observation variable given the previous echo state, which is compared against a threshold obtained by means of experimentation.

\subsection{Offline Training and Online Inference}
\cite{park_multimodal_2018} suggest using LSTM layers rather than dense layers in a VAE trained on nominal data in order to detect anomalies in robot-assisted feeding.
Rather than using a static prior, i.e.\ $\mathcal{N}(0, 1)$, they propose using a dynamic one, i.e.\ $\mathcal{N}(\mu_p, 1)$, which, according to the authors, introduces temporality into the prior distribution.
Furthermore, \cite{park_multimodal_2018} use a support-vector regressor (SVR) which is trained \lc{on the validation subset containing nominal data} to map latent distribution parameters $(\boldsymbol{\mu}_\mathbf{z}, \boldsymbol{\sigma}_\mathbf{z})$ to the resulting anomaly score.

\cite{su_robust_2019} propose a similar approach to \cite{park_multimodal_2018}, though with a few of modifications.
Firstly, the encoder and decoder are composed of gated recurrent unit (GRU) layers \cite{cho_properties_2014} rather than LSTM layers to minimise trainable parameters.
In addition to that, the previous hidden state in both the encoder and decoder is concatenated with the current input to explicitly represent temporal dependencies.
Lastly, the authors use planar normalising flow \cite{rezende_variational_2015} to model a non-Gaussian posterior through a predefined number of $K$ transformations to the latent vector.
The anomaly score used in this work is the reconstruction probability, i.e.\ the log probability of the reconstruction given the latent representation.
This is then compared with a threshold obtained using the peaks over threshold (POT) method, which works by fitting the tail of a modified generalised Pareto distribution with \lc{unlabelled} training data using maximum likelihood estimation.
This method is compared with grid-searching for the ideal $\mathrm{F_1}$ score and yields similar results.
Lastly, the authors propose a way to find the channel contributing to the anomaly to enable the interpretation of anomalous behaviour.

\cite{zhang_federated_2021} propose an approach to anomaly detection based on several convolutional GRU-based VAEs in a federated learning framework.
\lc{It is trained on nominal training data and the threshold is set as the maximum reconstruction error resulting from the validation data, which is nominal-only.}
Other than using convolutional GRU layers, the only contribution this paper makes revolves around federated learning and the authors show that the model on its own performs best outside the federated learning context.

\cite{li_mad-gan_2019} propose a way to detect anomalies using recurrent GAN.
Consisting of LSTM layers, both the discriminator and generator are used to compute an anomaly score.
During training, the generator uses random input vectors to generate credible time series\lc{, which the discriminator has to tell apart from nominal training data. During inference, however, }the input samples are mapped to a latent vector to serve as the input of the generator.
To obtain the anomaly score, a loss consisting of the weighted generator residuals and the weighted discriminator cross-entropy is used to compute an anomaly detection loss for each of the time-series windows.
\lc{It may be hard to apply to real-world problems as it not only requires a threshold but also the setting of the anomaly detection loss weight.}.

\lc{\cite{fan_luad_2023} introduce a VAE based on TCN layers and so-called hidden layers, though it is not specified what the underlying mechanism beneath the latter is. 
Given that the TCN layers are implemented ahead of the encoder and the decoder to capture the temporality of data it could be speculated that the hidden layers are simply dense layers.
Like \cite{su_robust_2019}, \cite{fan_luad_2023} use planar normalising flow \cite{rezende_variational_2015} to approximate a non-Gaussian prior.
During detection, anomaly scores for each dimension are ranked and weighted such that more anomalous channels are more prominent.
The threshold is obtained using the POT method.}

\lc{\cite{zhang_acvae_2024} propose a VAE regularised by two mechanisms: an adversarial one and a contrastive one.
The adversarial regularisation is sought by running the latent distribution parameters of the VAE through an additional model consisting of dense layers, which transforms the distribution parameters so that they represent an anomalous latent distribution.
Both latent distributions are then run through the decoder, yielding two different reconstructions.
The adversarial loss then is calculated by summing the Kullback–Leibler divergence between the two latent distributions and the error between the two reconstructions. 
Such a loss function causes the two distributions and reconstructions to be as close to each other as possible, which becomes relevant for the contrastive loss.
To obtain the contrastive loss, the two previously obtained reconstructions are re-encoded by the encoder which outputs another two latent distributions.
The contrastive loss is then calculated as a combination of the Kullback–Leibler divergence between the regular latent distribution and the re-encoded nominal latent distribution and of the Kullback–Leibler divergence between the regular latent distribution and the re-encoded anomalous latent distribution.
The anomaly score is defined as the mean-squared error and the threshold is obtained using kernel density estimation parameterised using training data.}

\subsection{Offline Training and Offline Inference but Online Capable}
Justified by the fact that real-world data does not always follow a standard Gaussian distribution, \cite{guo_multidimensional_2018} suggest using a Gaussian mixture prior to improved distribution mapping with multi-modal data.
In theory, this should not be required, as the job of the encoder is to map the input data that comes closest to a Gaussian prior, i.e.\ minimising the loss function.
The reconstruction probability is then used as the anomaly score, which is compared to a threshold based on a chosen percentile.
It is unclear whether this translates to tangible benefits in multivariate time series, since the data sets used found almost no application beyond this paper or are univariate.
No other publication proposes an approach that relies on a Gaussian mixture prior.

\cite{von_schleinitz_vasp_2021} apply anomaly detection in an attempt to make vehicle dynamics time-series prediction more robust to anomalies.
Anomalies in input data decrease the performance of prediction networks, therefore the authors implemented an LSTM-based VAE trained on unlabelled data to correct the input sample before it is used for prediction.
The anomaly score is given by a custom distance metric, similar to the z-score, which is compared with a threshold obtained from a grid search \lc{using labelled validation data}.

\cite{li_anomaly_2021} propose a GRU-based VAE with a custom prior acting as smoothing regularisation.
This forces the VAE to feature smooth transitions in distribution parameters along the time axis, which increases robustness \lc{when trained with unlabelled data containing both nominal and anomalous sequences}.
Like in most publications, the authors opted for the reconstruction probability as the anomaly score.

\cite{choi_multivariate_2022} create an LSTM-based VAE \lc{trained on nominal data} which features a 1D convolutional layer before the encoder to act as a feature extractor.
The anomaly score is taken as the Euclidean distance, which is compared to a threshold \lc{obtained in different ways depending on the data set.}

Autoencoders are also successfully combined with GANs.
Here, the generator input is a sample from a latent distribution obtained from the encoder, whereas in classical GANs the input to the generator is a random noise vector sampled from a distribution.

The first publication that proposes adding an encoder before the generator is by \cite{zhou_beatgan_2019}, where the discriminator is presented with real data and reconstructed data from a VAE rather than generated data from noise.
Only the autoencoder part is used for anomaly detection, as the anomaly score is calculated using the Euclidean distance between input and reconstruction.
The discriminator only plays a role in regularisation during training.

\cite{choi_gan-based_2020} employ a 2D CNN GAN which features an autoencoder-based generator in an effort to detect anomalies in power plant data.
Here, the input sequences are processed into Euclidean distance matrices \cite{lele_euclidean_1991} in order to increase robustness when faced with noise.
The anomaly score is then computed using the sum of the reconstruction error, given by the L1 norm, and the weighted feature loss, obtained using the output of an intermediate layer.
Here, the L2 norm between the output given the real sample and the output given the generated sample is taken as the feature loss.
The anomaly score is then compared to a threshold defined by a domain expert.

\cite{geiger_tadgan_2020} introduce an encoder to the GAN architecture as well as a second discriminator that measures how well the current latent distribution is mapped.
To prevent mode collapse, the authors opted to use the Wasserstein loss, as well as the L2 norm between original and reconstructed samples.
The anomaly score consists of a reconstruction component and a discriminator component.
To ensure a comparable scale on both components, they are standardised.
To combine the components sensibly, the authors propose to either use them as a weighted sum or as a weighted product.
They suggest calculating the reconstruction component in three different ways: using a simple difference, an area difference and the dynamic time warping distance, hence six different variations of the anomaly score are tested.
The anomaly score is then compared with a threshold, which is calculated as 4 standard deviations from the mean of the current window.

\cite{zhu_novel_2019} use a GAN composed of a CNN-based generator and a CNN and LSTM-based discriminator to detect anomalies\lc{, after being trained on nominal data.}
The anomaly score is computed using a weighted sum of generator and discriminator loss\lc{ which is then compared to a manually set threshold.}

\cite{sun_time_2019} propose a GAN composed of a dense generator and a 1D CNN discriminator to detect anomalies in sensor data from commercial vehicles, as well as predictive maintenance.
A model is trained for each channel, which is why a dense generator is applicable.
No anomaly score and threshold are required, since the approach uses the discriminator to detect the anomalies directly.

\section{Transformer Models}
\label{sec:transformer_modelling}
Transformers, first introduced by \cite{vaswani_attention_2017}, have been on the rise in the machine learning research landscape, thanks to significant advances enabled in natural language processing by the GPT-series of models \cite{radford_improving_2018, radford_language_2019, brown_language_2020, openai_gpt-4_2023}.
Transformer models have started to find their place in time-series anomaly detection, though by far to a lesser extent than the previously discussed model families.
The original transformer shows some resemblance to encoder-decoder architectures, though with a variety of improvements.
First, it does not use recurrent but rather feed-forward layers to process input information in an effort to accelerate training through parallelisation, which is otherwise a sequential operation with LSTM and GRU layers.
The temporal order of input data is maintained through the positional encoding of the inputs before entering the model.
Furthermore, transformers use an evolution of the attention mechanism, called multi-headed attention, which allows the model to attend to information more effectively.
A typical transformer block contains a feed-forward layer as well as a multi-headed attention layer, transformers can consist of several encoder and decoder blocks.
An overview of all the transformer-based modelling publications discussed in this survey is presented in Table \ref{tab:summary_tran}.
\begin{table}[h!]
\centering
\begin{tabular}{lcccc}
\hline
Publications                      & OT & OI         & DwL        & EM                        \\ \hline\hline
Zhang et al. \cite{zhang_unsupervised_2021-1} &    & \checkmark & \checkmark & $F_1$                     \\
Tuli et al. \cite{tuli_tranad_2022}          &    & \checkmark & \checkmark & $F_1$, $A_\text{ROC}$     \\
Doshi et al. \cite{doshi_reward_2022}         &    & \checkmark &            & $A_{P_\text{SA}\tilde D}$ \\
Yu et al. \cite{yu_dtaad_2024}             &    & \checkmark & \checkmark & $F_1$, $A_\text{ROC}$     \\
Xu et al. \cite{xu_anomaly_2022}           &    &            &            & $F_1$, $A_\text{ROC}$     \\
Chen et al. \cite{chen_learning_2021}        &    &            & \checkmark & $F_1$                     \\
Wang et al. \cite{wang_variational_2022}     &    &            &            & $F_1$                     \\ \hline
\end{tabular}
\caption{Summary of all publications surveyed in the area of transformer-based time-series anomaly detection. The key is as follows,  \lc{OT: online training, OI: online inference, DwL: detection without labels, EM: evaluation metrics, CM: confusion matrix.}}
\label{tab:summary_tran}
\end{table}

\subsection{Online Training and Online Inference}
Like with Subsection \ref{subsec:ononrec}, no transformer that trains and infers in an online manner exists, though, in theory, such approaches could work similarly to the solutions discussed in Subsection \ref{subsec:ononpred}.

\subsection{Offline Training and Online Inference}
\cite{zhang_unsupervised_2021-1} first combine VAEs with transformers (TransAnomaly) to detect anomalies.
Here, the latent representation output by the transformer encoder is a distribution that is sampled from to reconstruct the input signal.
The anomaly score is taken as the reconstruction probability, which is then compared with a threshold obtained with the POT method.

\cite{tuli_tranad_2022} propose using a pair of transformers which interface with each other to obtain two reconstructions.
The first receives input windows as the input, whereas the second receives a concatenation between the entire time series and a focus score.
The focus score is an error metric that describes the deviation between the input and reconstruction from the first transformer.
Training is done in two stages, where, in the first stage, the model attempts to minimise the reconstruction error (L2 norm) between the input window and the respective decoder reconstruction.
In the second stage, training becomes adversarial, the second decoder attempts to maximise the reconstruction error between the input and second decoder reconstruction, whereas the first decoder attempts to minimise it.
The anomaly score is then taken as the equal sum of the reconstruction errors obtained from both decoders.
The threshold is obtained using the POT method, like \cite{zhang_unsupervised_2021-1}.

Beyond proposing alternate metrics more suitable for evaluating the detection of sub-sequence anomalies, \cite{doshi_reward_2022} propose a novel transformer-based approach which is largely based on a so-called ProbSparse attention mechanism stemming from the informer model \cite{zhou_informer_2021}.
The goal of multi-head ProbSparse self-attention is to reduce computational complexity compared to regular multi-head self-attention \cite{vaswani_attention_2017}.
To perform anomaly detection, a k-nearest neighbours classifier is first trained on a nominal training subset split into two subsets.
During testing, the anomaly score for a given time step is calculated by subtracting a baseline based on a significance level from the difference between the distance between the current time step and the second subset used to train the k-nearest neighbours classifier.
The anomaly score then accumulates over time and leads to a raised flag if it exceeds a threshold for a specified minimum amount of time.

\lc{\cite{yu_dtaad_2024} proposes a model ensemble consisting of two transformers each with a TCN layer as feature extractors.
One of the TCNs has its receptive field configured narrowly for local information whereas the other in a wider manner for global information.
The transformer with the global TCN is fed with both input data, as well as the output from the transformer with the local TCN.
The anomaly score is denoted as the reconstruction error, with the threshold being obtained using the POT method.}

\subsection{Offline Training and Offline Inference but Online Capable}
\cite{xu_anomaly_2022} lay out an architectural change to regular transformers in the form of anomaly attention blocks which are used instead of self-attention blocks.
These anomaly attention blocks feature another parallel pipeline which takes in a fourth parameter to model the prior association, along with the query, key, and value computations found in a regular attention mechanism.
This is done in an effort to model the prior association and the series association at the same time, which regular transformers cannot.
The anomaly score is calculated using a product of the reconstruction error and the association discrepancy.

\cite{chen_learning_2021} add a context encoding block before the encoder block in a transformer to better model multivariate input dependencies through graph networks.
Anomalies are then detected using the L2 norm held up against a threshold obtained through $\mathrm{F_1}$ score-maximising grid search.

\cite{wang_variational_2022} take a similar approach to \cite{zhang_unsupervised_2021-1} by combining transformer models with VAEs.
Here, the authors add a feature extraction component which upsamples input data to reduce the sensitivity of the model when faced with corrupted data.
In addition to that, the authors do not use the reconstruction probability to detect anomalies, but rather by calculating the upper and lower error bounds for any given window.
These are calculated using the sum of the mean value and the product between the variance and a manually set parameter.

\section{Discussion}\label{sec:discussion}
Directly comparing the anomaly detection performance across the wide range of methods summarised in this survey is essentially impossible, unfortunately.
This is caused by a number of significant issues, of which we summarise the most important ones now.
Firstly, some authors presented their results solely as an $A_\text{ROC}$ figure, which, as specified in Subsection \ref{subsubsec:uncalibrated_metrics}, is unsuitable for anomaly detection evaluation.
Surprisingly, only \cite{li_anomaly_2021, kieu_outlier_2019} use the more appropriate $A_\text{PR}$ metric.
Secondly, very few papers use both calibrated and uncalibrated metrics to denote anomaly detection performance.
The two \lc{types of scores denote different aspects and} are incompatible with each other, hence comparing work that only uses one of the metrics is not possible.
Then, some publications use different types of $\mathrm{F}$ scores.
The most used version is the $\mathrm{F_1}$ score, where precision and recall are equally weighted, however, some publications use the $\mathrm{F_{0.5}}$ \cite{hundman_detecting_2018}, the $\mathrm{F_{0.1}}$ \cite{malhotra_long_2015, malhotra_lstm-based_2016, chauhan_anomaly_2015} or even the $\mathrm{F_{0.05}}$ score \cite{malhotra_lstm-based_2016} which favours precision over recall.
For obvious reasons, this makes a comparison of the approaches difficult, however, this is a rare pattern throughout the literature and exclusively appeared in older publications.
Even when considering the papers that included $F_1$ scores, the numbers are mostly incomparable.
Since \cite{xu_unsupervised_2018} proposed it, several papers started using point-adjusted metrics, which inflates evaluation metrics and makes the approaches look better than they actually are.
For more information on evaluation metrics and their flaws, refer to Subsection \ref{subsec:metrics}.
The lack of consensus regarding the evaluation of sub-sequences further prevents comparison.
\lc{Additionally, little attention is paid to discrete-sequence problems, which most proposed evaluation metrics are unsuitable for, especially because they assume that there is always a predicted positive sub-sequence.}

A large number of publications use solely proprietary data sets, which, while valid due to the flaws discussed present in public data sets (Subsection \ref{subsec:data_sets}), hinders direct comparison with other approaches and reproducibility.
\lc{Some public data sets do not provide pre-determined training and testing subsets, hence they are likely to be split differently in any publication using them, preventing comparison between approaches.
Furthermore, the data sets that do provide training and testing splits, are also vague as to what exactly the training subset is composed of. 
Not providing labels for each sequence in the training subset is consistent with the unsupervised problem setting, but information on the contamination level, i.e.\ whether anomalies are present in the subset, is not provided.}
Lastly, some publications only use a subset of a given data set, which further prevents a fair comparison, especially when performance metrics are provided as an average over all subsets.

Another observation made is that models that both detect and learn in an online manner rarely find applications. They only appeared a few times in the predictive modelling landscape, once in the generative modelling area and nowhere in the remaining sections. One possible explanation is that with increasing model complexity and parameter count, more and more data is required to properly model the data distribution, which is not available at inference time.

\lc{Often publications choose a threshold using labelled data, such as \cite{malhotra_long_2015, naito_anomaly_2021, suh_echo-state_2016, doshi_reward_2022} to name a few, which is rarely available in the real world. 
Sometimes, assumptions are made, such as using a given percentile of the anomaly score from validation data \cite{filonov_rnn-based_2017, guo_multidimensional_2018}, though justification is rarely provided.
Yet others \cite{su_robust_2019, zhang_unsupervised_2021-1, tuli_tranad_2022} use the POT method, an automatic threshold selection procedure, which still requires to be parameterised and hence requires some amount of labelled data.
Active learning could serve as an alternative as a domain expert can actively label windows or sequences which can be used to further improve the detection performance of a model initially trained in a semi-supervised or unsupervised manner.
This can be of benefit to the real-world application of the approaches discussed in this work, since querying and acquiring labels with time allows users to gauge the detection performance of the approach used, leading to more transparency in an otherwise unsupervised setting.
The main challenge is that models trained in an unsupervised manner are inherently incompatible with labelled data, hence enhancing the threshold choice with the labelled data offers a more straightforward alternative to tweaking model parameters. 
\cite{lundstrom_interactive_2023}, for example, proposes an active learning-based threshold setting procedure which carries the benefit of requiring especially little queries. 
First, an initial threshold is set using a method similar to  \cite{hundman_detecting_2018} but with the modification that the entire anomaly score sequence is used.
Then, predicted anomalous sub-sequences are clustered using four attributes: temporal distance to each other, the difference between maximum magnitude in each channel, the difference between maximum anomaly score for each channel and which channels are predicted anomalous.
When a sub-sequence is labelled, all other sub-sequences in the same cluster are assumed to have the same label, which reduces the querying budget required.
This method has a number of disadvantages, however. For a sub-sequence to be assigned to a cluster it needs to meet a number of manually set conditions, like how large the temporal distance can be and how large the difference between magnitudes and maximum anomaly scores can be.
Additionally, it can only be applied to methods which output an anomaly score per channel and it assumes an offline setting, i.e. all time steps of the testing sequence are available.
}
Going further, \cite{homayouni_autocorrelation-based_2020} enhance autoencoder structures using labelled data by modifying the loss function to allow for the ingestion of labelled data.
\lc{Both \cite{lundstrom_interactive_2023} and \cite{homayouni_autocorrelation-based_2020} assume a perfect domain expert which always labels correctly.
In the real world, that is rarely the case, hence the probability of mislabelling should be taken into account to yield robust methodology.}

Of the four model groups identified, transformer-based approaches have received the least attention so far, due to their relatively recent introduction in 2017.
The authors believe that the number of publications in this area is set to rise in the coming years.
Given the portfolio of publicly available data sets and their shortcomings, the application of transformer-based models still needs to be justified with a dataset that is too complex for simpler approaches.

\section{Real-world Applications}\label{sec:applications}
\lc{Generally, time-series anomaly detection can be applied to any use case involving a dynamic system, i.e.\ a system whose behaviour is a function of time. 
The approaches surveyed are used in the context of a myriad of applications.
They include:
\begin{itemize}
    \item Computing systems \cite{yen_causalconvlstm_2019, su_robust_2019}
    \item Automotive \cite{von_schleinitz_vasp_2021, sun_time_2019}
    \item Aerospace \cite{que_real-time_2019, hundman_detecting_2018}
    \item Healthcare \cite{thill_anomaly_2019, sakuma_area-efficient_2021, thill_temporal_2021, zhou_beatgan_2019, park_multimodal_2018}
    \item Manufacturing \cite{hsieh_unsupervised_2019, bayram_real_2021, lindemann_anomaly_2020, tayeh_attention-based_2022}
    \item Chemical processes \cite{filonov_multivariate_2016, filonov_rnn-based_2017,naito_anomaly_2021}
    \item Infrastructure \cite{zhang_deep_2019, li_mad-gan_2019, choi_gan-based_2020}
    \item Telecommunications \cite{audibert_usad_2020}
\end{itemize}
One key benefit of model-based approaches is that they require little to no domain knowledge since any relevant information is assumed to be present within training data. 
Therefore, while some approaches are presented in the context of a specific application, they are often generic and can be flexibly applied to other domains.
}
 
\section{Conclusion}\label{sec:conclusion}
This survey provides an overview of the model-based approaches used to detect anomalies in multivariate time series \lc{as well as of the data sets and evaluation metrics used to compare them and deeper accompanying analysis.}
\lc{Furthermore, it makes two novel contributions to the taxonomy of the research area, by making a distinction between online inference and online training and additionally between continuous-sequence and discrete-sequence anomaly detection in multivariate time series.}
The surveyed model-based approaches are categorised into four model groups, then further segmented into online training and online inference, offline training and online inference, and offline training and offline inference but online capable. 
\lc{For each model group, approaches are segmented by anomaly detection type as well as the metrics used to quantify anomaly detection performance.}
The publications analysed are then sorted by similarities in architecture and publishing date.
Therefore, this work provides the reader with a clear picture of the online time-series anomaly detection landscape, as well as potential research areas that future work could focus on and issues that hinder progress in said area.

Several research gaps are identified, some centred around unexplored model types and configurations, but also around the lack of consensus when it comes to benchmarking approaches.
To be able to properly quantify the progress made in this research area, consensus needs to be established throughout the time-series anomaly detection field.
\lc{Especially in the case of the data sets and evaluation metrics, several publications are made in an effort to advance the research area, though after thorough analysis, a number of problems remain unaddressed.}

\lc{Therefore, before the current research can be successfully applied by practitioners in real-world use cases, a number of conditions need to be fulfilled.
As highlighted above, the time-series anomaly detection field must find common ground in terms of the reproduction of results.
Without reliable results obtained from testing resembling real-world problems, it is impossible to know how well approaches actually perform.
Additionally, the metrics used to arrive at said results must be interpretable and reflect intuition.
When the groundwork has been set, tangible advances in the field can be made, which in turn will allow for the wider application of time-series anomaly detection approaches.
}

\section{Outlook}\label{sec:outlook}
\lc{Future work should first and foremost revolve around benchmarking, which consists of publicly available data sets and evaluation metrics that intuitively reflect detection performance.
In general, there is a need for a data set that not only is at least real-world-like but is also diverse and features anomalies that resemble realistic and hard-to-spot cases.
Future benchmarking data sets should provide clearly defined training and testing subsets and specify whether the training subset is apt for semi-supervised anomaly detection, i.e.\ consisting of nominal sequences only, or for unsupervised anomaly detection, i.e.\ contaminated with a small amount of anomalous data.
When it comes to evaluation metrics, this work proposes, as previously mentioned, a novel way to divide the time-series anomaly detection field: continuous-sequence and discrete-sequence anomaly detection. 
Future work should aim to provide a set of metrics that can be applied to both cases while maintaining interpretability and requiring no parameters.
Once these foundations are set, work needs to be dedicated to a definitive comparison between deep learning-based and traditional approaches. As mentioned by \cite{wu_current_2021}, so far there is no evidence that model-based approaches relying on deep learning are necessary to detect anomalies in the case of current public data sets.
Only then real and measurable advances be made in the time-series anomaly detection field.}

\lc{Aside from that, a few research gaps related to the modelling process are identified. 
The fast-moving field of natural language processing offers many approaches that could find application in time series modelling, given the similarity between the two fields.
Future work should monitor on advances in the said field to potentially apply them to time-series anomaly detection.
In addition to that, attention should be paid to how humans interface with the models, i.e.\ how a human can enhance anomaly detection approaches.}
\printbibliography %Prints bibliography

\end{document}